\documentclass[preprint,12pt]{elsarticle}

\usepackage{graphicx}
\usepackage{multirow}
\usepackage{amsmath,amssymb,amsfonts}
\usepackage{amsthm}
\usepackage{mathrsfs}
\usepackage{xcolor}
\usepackage{textcomp}
\usepackage{booktabs}
\usepackage{array}
\usepackage{longtable}
\usepackage{rotating}
\usepackage{ragged2e}
\usepackage{caption}
\usepackage{hyperref}
\usepackage{lineno}
\usepackage{adjustbox}
\usepackage{algorithm}
\usepackage{algorithmicx}
\usepackage{algpseudocode}

\theoremstyle{plain}

\theoremstyle{definition}

\theoremstyle{remark}

\modulolinenumbers[5]

\begin{document}

\begin{frontmatter}

\title{Large AI Models in Dental Healthcare: From General-Purpose Systems to Domain-Specific Foundation Models}

\author[uaeu]{Sema Helali}
\ead{700049724@uaeu.ac.ae}

\author[uos]{Lina Abu Nada}
\ead{lnada@sharjah.ac.ae}

\author[uos]{Sausan Al Kawas}
\ead{sausan@sharjah.ac.ae}

\author[wcm]{Alaa Abd-Alrazaq}
\ead{aaa4027@qatar-med.cornell.edu}

\author[mcgill]{Faleh Tamimi}
\ead{faleh.tamimimarino@mcgill.ca}

\author[uaeu,mcgill]{Rafat Damseh\corref{cor1}}
\ead{rdamseh@uaeu.ac.ae; rafat.damseh@mcgill.ca}

\cortext[cor1]{Corresponding author}

\affiliation[uaeu]{
  organization={Department of Computer Science and Software Engineering,
                College of Information Technology,
                United Arab Emirates University},
  city={Al Ain},
  postcode={15551},
  country={UAE}
}

\affiliation[uos]{
  organization={Department of Oral and Craniofacial Health Sciences,
                College of Dental Medicine,
                University of Sharjah},
  city={Sharjah},
  country={UAE}
}

\affiliation[wcm]{
  organization={Weill Cornell Medicine-Qatar},
  city={Doha},
  state={Education City},
  country={Qatar}
}

\affiliation[mcgill]{
  organization={Faculty of Dental Medicine and Oral Health Sciences, McGill University},
  city={Montreal},
  state={Quebec},
  country={Canada}
}

\begin{abstract}

\normalsize
\textbf{Background.}
Oral diseases affect nearly 3.5 billion people worldwide, yet the clinical role of large-scale AI models in dentistry remains unclear. Recent work spans language-generative models, vision foundation models, and dental-specific foundation models, but no unified review has compared their clinical roles.

\normalsize
\textbf{Methods.}
Following PRISMA-ScR guidelines, we searched PubMed, Google Scholar, Scopus, and arXiv. Two reviewers screened studies, and 97 studies from 2020 to 2026 met the inclusion criteria. We propose a two-dimensional framework classifying models by architecture and dental specialization, then analyze performance, methods, and limitations across clinical, educational, and imaging applications.

\normalsize
\textbf{Results.}
Language-generative models perform well on text-based tasks such as clinical reasoning, exams, and patient communication, but are less reliable for image-dependent diagnosis. Adapted vision foundation models, especially SAM and CLIP variants, perform strongly in tooth segmentation and lesion detection. Dental-specific models, including DentVFM, DentVLM, and OralGPT, show the strongest results on complex multimodal tasks. Overall, integrated pipelines outperform single-model approaches.

\normalsize
\textbf{Conclusions.}
General-purpose and dental-specific models are complementary rather than competing. The strongest dental AI systems combine both within structured pipelines. Safe clinical deployment still requires better hallucination control, more annotated dental data, and standardized evaluation benchmarks.
\end{abstract}


\end{frontmatter}


\section{Introduction}

\begin{figure}[ht]
\centering
\includegraphics[width=\linewidth]{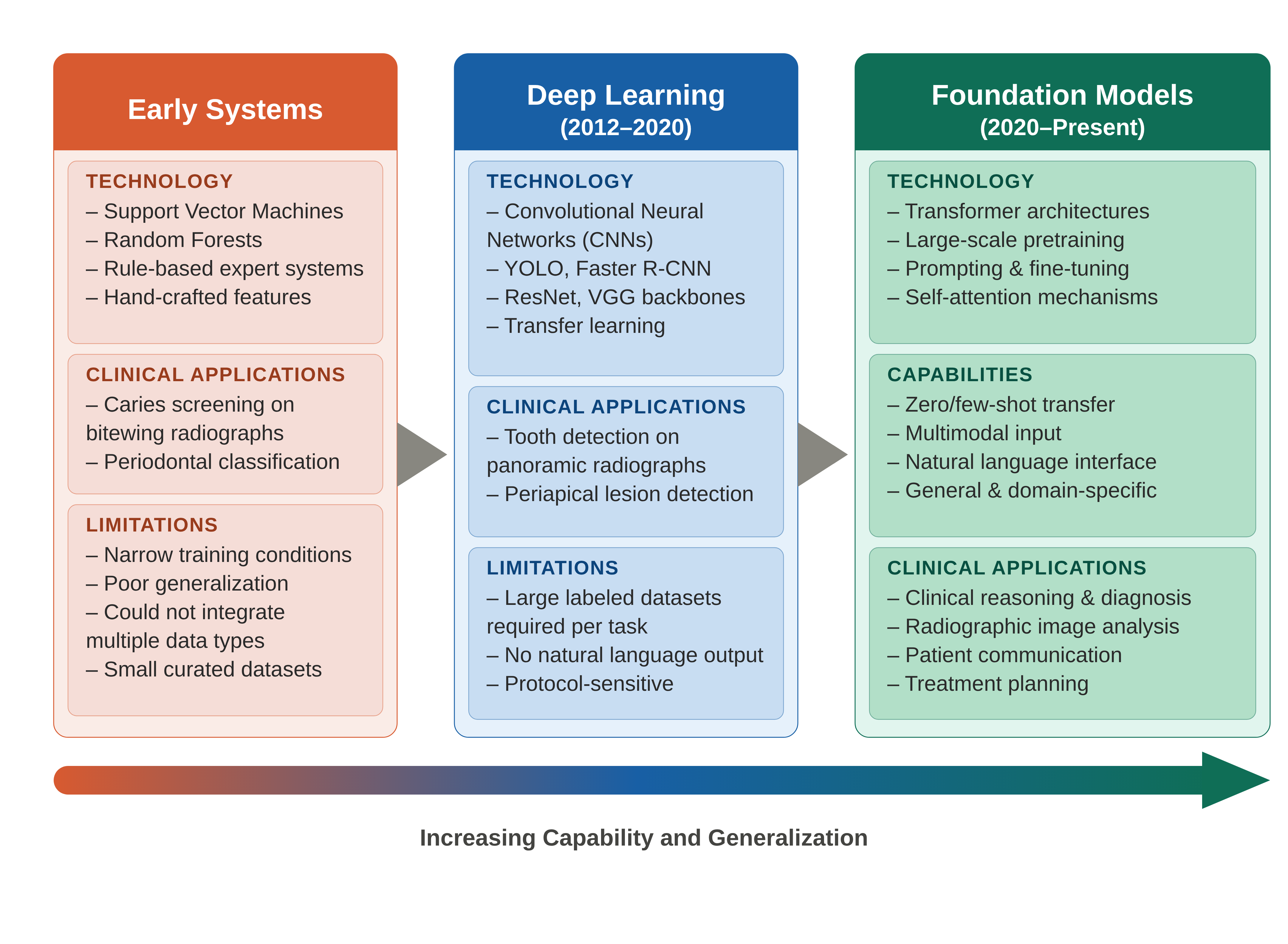}
\caption{The evolution of AI technology in dental practice, illustrating
the progression from early rule-based and classical machine learning systems
through the deep learning era to modern foundation models. Advances in model
architecture, data utilization, and learning strategies have progressively
increased generalizability, multimodal capability, and the range of clinical
applications.}
\label{fig:fig1}
\end{figure}

Oral diseases affect nearly 3.5 billion people, close to half the world's
population, which makes them the most widespread non-communicable diseases
globally~\cite{WHO2022OralHealth}. A burden of this scale has driven decades of
effort to bring artificial intelligence (AI) into dental practice, and that
effort has passed through three generations, each defined by what its models
could and could not do. Figure~\ref{fig:fig1} traces this progression.

The first two generations preceded foundation models and shared the same ceiling.
The first generation relied on hand-crafted features fed into support vector
machines, decision trees, and rule-based expert systems. These worked well on
narrow, well-defined problems, such as caries screening on bitewing radiographs
or rule-based classification of periodontal status, but generalized poorly beyond
the conditions they were built for~\cite{Virupaiah2020AnalysisOI,article}. The
second generation, built on convolutional neural networks (CNNs), raised the
performance ceiling considerably and reached specialist-level results on tasks
such as tooth detection and numbering, periapical lesion identification, and
restoration segmentation~\cite{tuzoff2019tooth,lee2021deep,li2021detection}. Even
so, both carried the same structural limits: each application needed its own large
labeled dataset, performance dropped when imaging hardware or protocols changed,
and the models produced no natural-language output that a clinician could use
directly.

Foundation models changed these terms. Trained at scale on large, heterogeneous
corpora, transformer-based models support zero-shot and few-shot transfer, handle
several modalities within one architecture, and interact through natural
language~\cite{vaswani2017attention,brown2020language}. Rather than being trained
from scratch for each task, they are pretrained once and then directed toward new
tasks through prompting or lightweight fine-tuning. In dentistry this made
tractable a range of applications that earlier architectures could not handle,
including clinical reasoning, imaging analysis, patient communication, and
educational assessment. The work that followed has taken two forms:
general-purpose models applied to dental tasks through prompting and fine-tuning,
and domain-specific models pretrained on curated dental
corpora~\cite{Fanelli2025,huang2025dentvfm}.

Despite the volume of work now published, the field still lacks a coherent
comparative account of these model classes. Studies have largely addressed one
category at a time: language models evaluated on clinical question answering,
vision foundation models adapted for radiographic segmentation, and
dental-specific architectures benchmarked on narrow imaging
tasks~\cite{Zhang2025EASAM,meng2025dentvlm,zhang2025oralgpt}. How the categories
relate to one another, where they are complementary, where one outperforms
another, and what their shared limitations mean for safe deployment, has not been
examined within a single framework. Without that perspective, choosing the right
model class for a clinical application, or deciding when to combine approaches,
cannot be done on principled grounds.

This review addresses that gap. Following PRISMA-ScR
guidelines~\cite{Page2021PRISMA}, we systematically searched four major databases
and analyzed 97 studies published between 2020 and 2026, covering general-purpose
LLMs, vision-language models VLMs, and dental-specific foundation models across
clinical, educational, and imaging applications. The review has four aims: to
synthesize this literature reproducibly and transparently; to propose a
classification scheme that organizes these model classes by architectural
paradigm and degree of dental specialization; to critically analyze and compare
their performance, methods, and limitations across application domains; and to
identify the open challenges that must be resolved before these systems can be
deployed autonomously in clinical practice.

The contributions of this work are as follows:
{\normalsize
\begin{itemize}
    \item \textbf{Systematic literature synthesis:} A reproducible PRISMA-ScR search and selection of 97 studies from four major databases, covering large AI model applications in dental healthcare from 2020 to 2026.

    \item \textbf{Classification framework:} A two-dimensional taxonomy classifying large dental AI models by architecture and dental specialization, enabling structured comparison across model classes.

    \item \textbf{Critical performance analysis:} A critical review of model performance, methods, and limitations across model categories and clinical domains, assessing both quality and generalizability.

    \item \textbf{Cross-model complementarity:} Evidence that integrated multi-model pipelines combining general-purpose and dental-specific models outperform single-model approaches across several task categories.

    \item \textbf{Challenges and future directions:} A structured discussion of deployment barriers, including hallucination, limited annotated dental data, and missing standardized benchmarks, with recommendations to address them.
\end{itemize}
}

\section{Search Strategy and Study Selection}\label{sec4}

This review sought to identify studies published between 2020 and early 2026
that developed, evaluated, or applied large-scale AI models in dental healthcare
contexts. The study selection process followed the PRISMA-ScR
framework~\cite{Page2021PRISMA}, covering four stages: identification, screening,
eligibility, and inclusion. A list of abbreviations used throughout this
review is provided in Table~\ref{tab:abbreviations}.

\textbf{Identification.}
A comprehensive literature search was conducted across four databases: PubMed,
Google Scholar, Scopus, and arXiv. These databases were selected to provide
complementary coverage: PubMed captures peer-reviewed clinical and biomedical
literature; arXiv ensures inclusion of recent AI preprints not yet formally
published; and Scopus and Google Scholar provide broad interdisciplinary reach
across engineering, computer science, and health informatics. The search strategy
focused on three main domains: large language models in dental applications,
multimodal and vision-language models in dental healthcare, and dental-specific
foundation models.

Search terms were combined using Boolean operators (AND, OR) across three
categories to capture the full range of relevant work. AI model terms included
"large language model," "GPT," "ChatGPT," "Claude," "Gemini," "foundation model,"
"vision-language model," "multimodal model," "SAM," "Segment Anything Model,"
"transformer," "BERT," and "vision transformer." Dental-related terms included
"dental," "dentistry," "oral health," "tooth," "teeth," "periodontal,"
"orthodontic," "endodontic," "oral surgery," "maxillofacial," "caries,"
"panoramic radiograph," "CBCT," and "dental imaging." Application-related terms
included "diagnosis," "detection," "segmentation," "classification," "clinical
decision support," "treatment planning," "education," and "patient communication."

\textbf{Screening.}
The initial search identified 1,129 records across the four databases. After
removing duplicates, 823 papers remained. Title and abstract screening was
conducted independently by two reviewers (S.H. and R.D.) using predefined
inclusion and exclusion criteria. Disagreements were resolved through discussion
and consensus. Studies were considered eligible if they investigated, developed,
evaluated, or applied large-scale foundation models in dental or oral health
contexts. Eligible study types included original research, technical reports,
validation studies, and comparative studies published between 2020 and 2026 and
written in English. Studies were excluded if they were review papers, pilot
studies, unrelated to dentistry, or focused solely on traditional machine learning
or conventional deep learning without a foundation model component. After this
stage, 214 papers remained.

\textbf{Eligibility.}
The remaining 214 studies underwent independent full-text screening by both
reviewers (S.H. and R.D.). After attempting full-text retrieval, 30 reports could not be accessed due to unavailable or restricted full texts, leaving 184 studies for assessment. Papers were excluded if they reported no assessable or sufficiently rigorous evaluation of model performance (n = 25), lacked a clear and direct dental application (n = 22), or were redundant or overlapping with already included work (n = 34). Studies with very small sample sizes that did not allow for meaningful performance assessment (n = 6) were also excluded. Any disagreements at this stage were resolved through discussion, with a third reviewer consulted where consensus could not be reached.

\textbf{Data extraction.}
Data were extracted from each included study using a standardized charting form
covering: study aim, model type and architecture, dental application domain,
dataset characteristics, evaluation metrics, key results, and reported
limitations. Extraction was performed by S.H. and verified by R.D. Discrepancies
were resolved through discussion.

\textbf{Quality appraisal.}
Formal quality appraisal of individual studies was not conducted. In accordance
with JBI methodology for scoping reviews, which aim to map the available evidence
rather than assess its quality, appraisal was considered outside the scope of
this review. Methodological limitations of individual studies are instead noted
narratively within the relevant results sections.

\begin{figure}[!t]
    \centering
    \includegraphics[width=\linewidth]{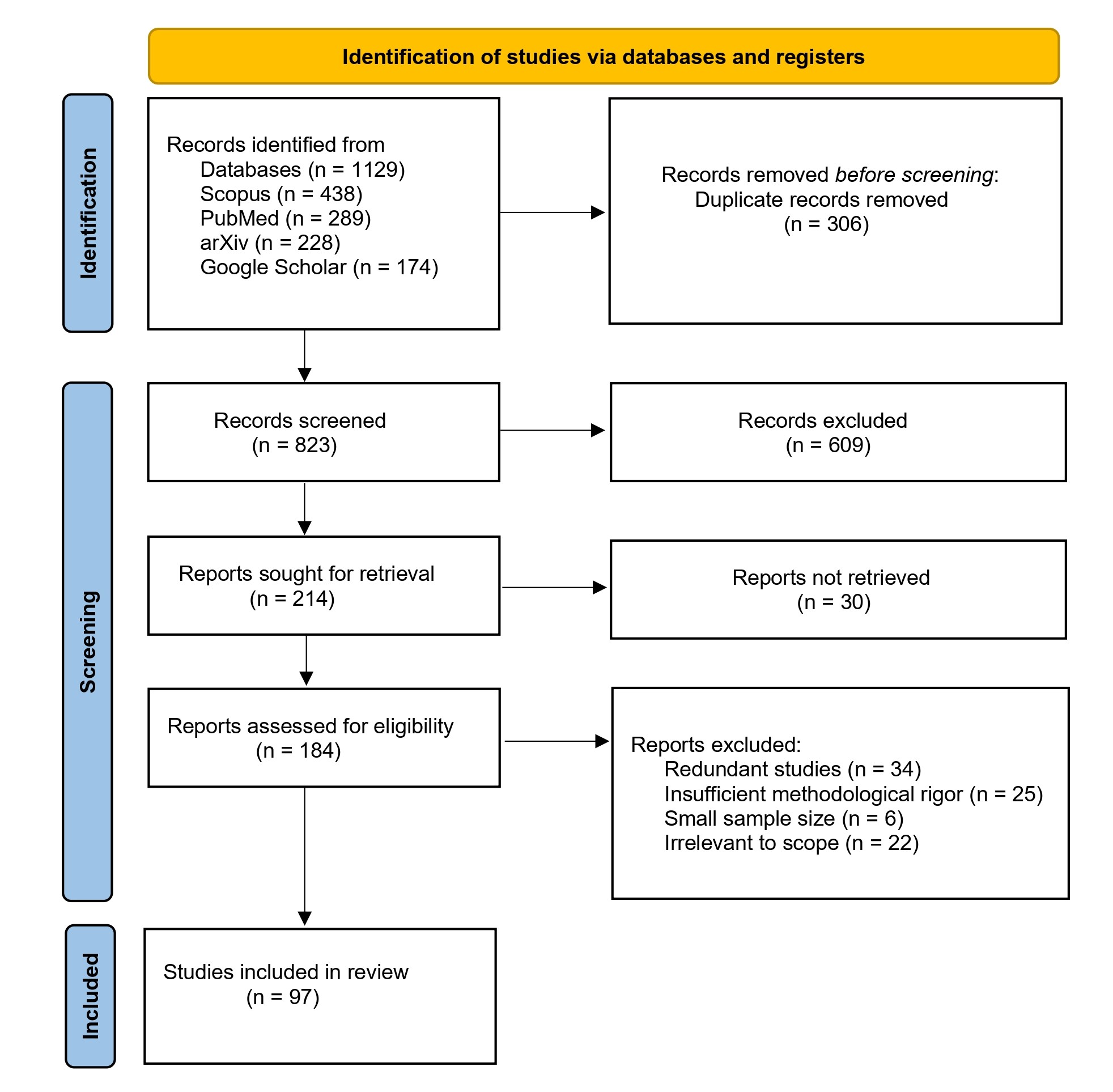}
    \caption{PRISMA 2020 flow diagram summarizing the systematic process of
    study identification, screening, eligibility assessment, and inclusion.
    Records were retrieved from four databases, duplicates were removed, and
    remaining studies were independently screened by two reviewers (S.H. and
    R.D.) based on predefined inclusion and exclusion criteria.}
    \label{fig:prisma}
\end{figure}

\textbf{Included studies.}
After applying these criteria, 97 papers were included in the final analysis,
as illustrated in the PRISMA flow diagram in Figure~\ref{fig:prisma}.

The classification framework used to organize these 97 studies by architectural
paradigm and degree of dental specialization is described in the following section.

\section{Conceptual Framework of Large AI Models in Dental Healthcare}\label{sec:background}

Classifying large AI models in dental healthcare is not straightforward. These systems differ in their inputs, outputs, and links to dental knowledge.

Rather than building an exhaustive taxonomy, this review organizes the 97 studies identified in Section~\ref{sec4} along two main axes. The first is architectural paradigm, which distinguishes models that generate language from models that produce structured visual representations. The second is degree of dental specialization, which captures how far a model has been adapted from general-purpose use toward clinical dental applications.

\begin{figure}[ht]
    \centering
    \includegraphics[width=\linewidth]{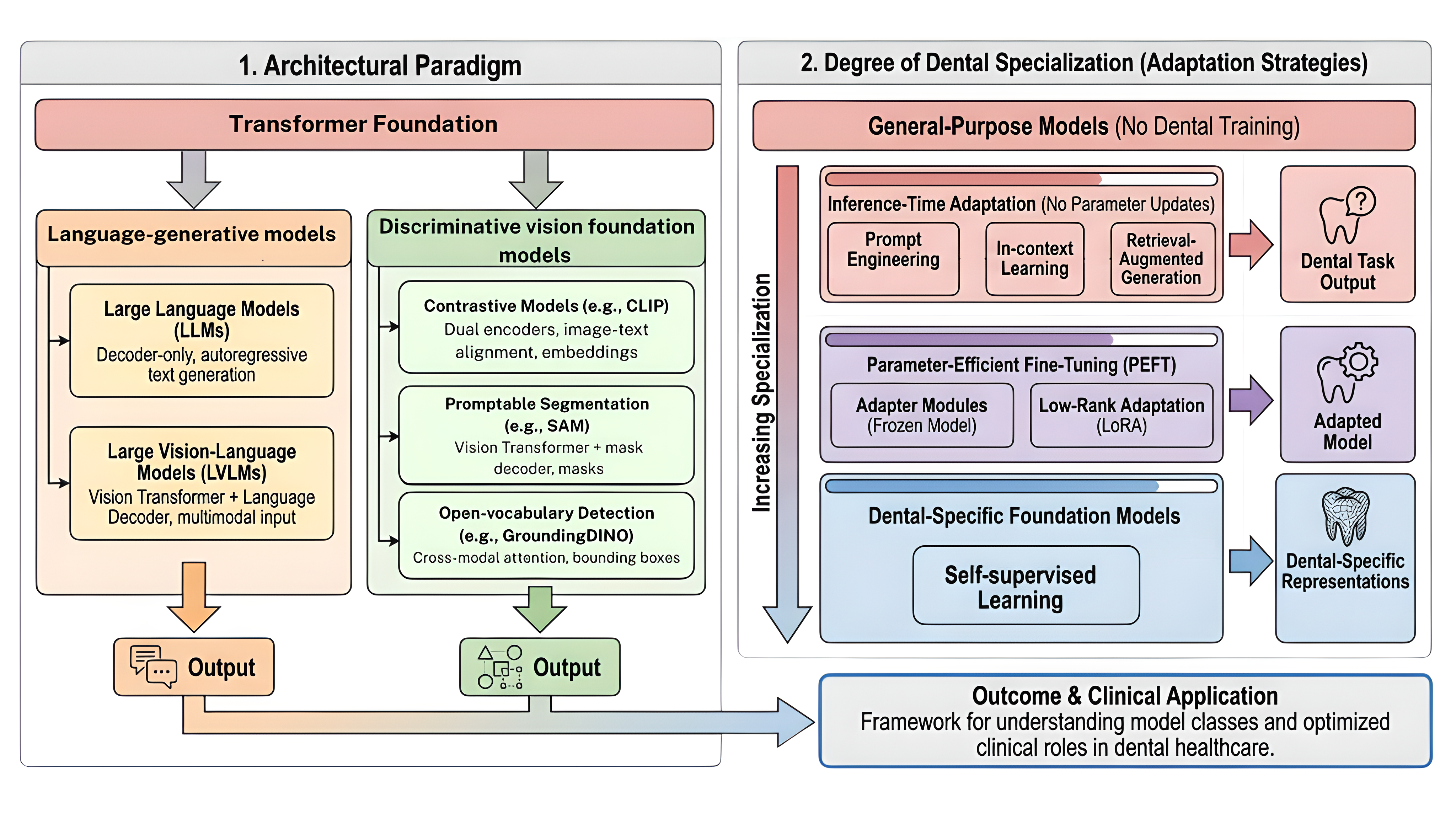}
    \caption{Large AI models in dental healthcare organized by architectural
    paradigm (left) and degree of dental specialization (right). The left panel
    distinguishes language-generative models, which produce language, from
    discriminative vision foundation models, which produce structured visual outputs such
    as embeddings, masks, or bounding boxes. The right panel shows how
    general-purpose models can be progressively adapted to dental tasks through
    inference-time strategies, parameter-efficient fine-tuning, or full
    domain-specific pretraining.}
    \label{fig:notax}
\end{figure}

Figure~\ref{fig:notax} summarizes the relationship between these model categories and their clinical roles.

\subsection{Architectural Paradigm}

All models reviewed here are built on the transformer
architecture~\cite{vaswani2017attention}, but they split into two families based
on what they produce.

\normalsize
\textbf{Language-generative models} produce language as output. They use
decoder-only transformers trained autoregressively on text: the model learns
to predict the next token from all preceding context~\cite{brown2020language},
and at sufficient scale this produces systems capable of reasoning, instruction
following, and in-context generalization. Vision-language models extend this by
coupling a visual encoder (typically a Vision Transformer that converts an
image into a sequence of patch embeddings~\cite{dosovitskiy2020vit}) to a
language decoder through a learned projection
layer~\cite{liu2023visual,hurst2024gpt}. The result accepts both images and text
as input while generating language as output, which makes these models suitable
for visual question answering, image-grounded reasoning, and radiographic report
generation. The defining property of this family, whether text-only or
image-conditioned, is that they generate language. Although vision-language
models such as GPT-4o and Gemini accept image input, they remain within the
language-generative family because their primary output is generated language;
the addition of a visual encoder does not alter the generative nature of the
underlying decoder.

\normalsize
\textbf{Discriminative vision foundation models} learn visual representations
without a language generation objective. Contrastive models such as CLIP align
image and text encoders in a shared embedding space~\cite{radford2021learning},
enabling zero-shot classification through similarity comparison. Promptable
segmentation models such as SAM use a Vision Transformer encoder paired with a
lightweight mask decoder, accepting point or bounding-box prompts to generate
segmentation masks across image domains~\cite{kirillov2023segment}. Open-vocabulary
detection models such as GroundingDINO incorporate language into the detection
pipeline via cross-modal attention, allowing object localization from arbitrary
text queries rather than fixed category
sets~\cite{liu2024groundingdino}. Architecturally diverse as these models are,
they share one property that separates them from the generative family: their
outputs are structured visual representations: embeddings, masks, or bounding
boxes, not language.

This distinction matters clinically. Language-generative models are the natural fit for
tasks that require explanation, dialogue, or text output: clinical question
answering, patient communication, and report drafting. Discriminative vision foundation models are
better suited to tasks that require spatial precision: tooth segmentation,
lesion localization, and anatomical landmark detection. Many effective dental AI
pipelines combine both, using a vision model to extract structured findings and
a generative model to interpret and communicate them.
\subsection{Degree of Dental Specialization}

General-purpose models are not trained on dental data, so using them clinically
requires some form of adaptation. The available strategies fall into three tiers,
ordered by how much of the model they change. The right tier depends on how much
labeled dental data exists, how much compute is available, and how large the gap
is between the model's general pretraining and the target dental task.

\normalsize
\textbf{Inference-time adaptation} leaves the model's weights untouched and acts
only on what is fed into it. Prompt engineering writes the input deliberately,
through instructions, formatting, or role framing, to steer the output.
Chain-of-thought (CoT) prompting is one such technique: the model is asked to work
through the intermediate steps before committing to an answer, which raises
accuracy on problems that take several reasoning steps~\cite{wei2022cot}.
In-context learning (ICL) places a few solved examples in the prompt so the model
infers the pattern and applies it to a new case, learning the task from the
examples alone without any weight update~\cite{brown2020language}.
Retrieval-augmented generation (RAG) goes a step further: before answering, it
retrieves relevant passages from an external knowledge base and inserts them into
the prompt, so the response is grounded in those sources rather than the model's
fixed memory. This keeps clinical knowledge current and reduces hallucination, but
needs a retrieval system and a maintained knowledge base that prompting does
not~\cite{lewis2020rag}. None of these require annotated dental data, which makes
this the default entry point.

\normalsize
\textbf{Parameter-efficient fine-tuning (PEFT)} adapts a large pretrained model by freezing most of its weights and training only a small set of additional parameters\cite{houlsby2019adapters}. This reduces computational cost and lowers the risk of overfitting when dental datasets are limited. In this review, PEFT mainly refers to LoRA and QLoRA, which learn compact low-rank updates to existing weights rather than retraining the full model~\cite{hu2021lora,dettmers2023qlora}. These methods preserve most of the model’s general knowledge while steering it toward dental data. This is especially useful for vision foundation models, where PEFT can help bridge the gap between natural-image pretraining and dental radiographic appearance.

\normalsize
\textbf{Full dental specialization} rebuilds the training signal itself rather than
adapting a fixed model. Self-supervised pretraining, such as masked image modeling,
hides parts of unlabeled dental radiographs and trains the model to reconstruct
them, so it learns dental structure directly from images without manual
labels~\cite{he2021mae}. Instruction tuning then trains the model on many
expert-written question-answer pairs so it follows clinical task
instructions~\cite{ouyang2022instructgpt}. Knowledge learned this way is held in
the weights rather than supplied at inference, which gives a consistent advantage
on tasks needing stable anatomical understanding across imaging protocols and
patients, at the cost of large curated datasets and heavy compute.

No tier is best in general; each balances cost, flexibility, and performance. Inference-time strategies are low-cost and easy to update as guidelines change, while fully specialized models require more data and maintenance but can perform better when dental-specific visual or linguistic knowledge is essential. Thus, the best approach is to use the lightest tier that meets the task’s needs. In practice, tiers are often combined, such as LoRA with RAG or dental vision models with general language decoders, and the strongest systems reviewed here follow this integrated pattern.

The two axes are not fully independent. From-scratch dental pretraining appears
almost entirely within the discriminative vision family, because large curated
dental imaging datasets exist whereas comparable dental text corpora do not.
Language-generative models are therefore specialized only by fine-tuning general
models, while discriminative vision models span the full range from zero-shot use
to from-scratch dental pretraining. This asymmetry is itself a finding about the
current state of the field.

\section{General Generative Models in Dentistry}\label{sec:gllm}
These models generate language, making them useful for dental tasks involving clinical reasoning, exams, dialogue, and reports. Evidence is grouped into clinical
reasoning and diagnostic support, dental education, and patient communication. Across all areas, they perform well on text-based tasks but poorly when image interpretation is required.

\normalsize
\subsection{Clinical Reasoning and Diagnostic Support}\label{sec:clinreasoning}
Across clinical reasoning and diagnostic tasks, language-generative models show considerable potential, but performance varies by task type. Results are strongest for text-based reasoning, less consistent for image interpretation, and more variable for clinical decision-making, as reflected in the studies summarized in Tables~\ref{tab:llm_dental_1a}.

\subsubsection{Text-Based Clinical Reasoning}
Text-based clinical reasoning represents one of the strongest application areas for language-generative models in dentistry. However, it is not a fixed capability that a model either has or lacks. Across the studies reviewed here, performance depends on three variables that are often confused with each other: how the case itself is described, what format the question takes, and which specific model or system is used to answer it.

The clearest demonstration of the first variable comes from oral pathology, where GPT-4 approached specialist-level performance on oral and maxillofacial disease cases while substantially outperforming GPT-3.5~\cite{tomo2024chatgpt}, yet that same accuracy fell from 80.18\% to 61.80\% once non-standardized descriptions were used. This shows the decline was caused by how the case was written, not by a change in the model's underlying knowledge. This suggests that reliable performance depends heavily on the quality and structure of clinical information provided to the model, a dependency that real clinical notes, written inconsistently across providers, do not reliably satisfy.

The second variable, question format, becomes clear once results are compared across different types of questions. Open-ended endodontic question answering reached 82.5\% accuracy for GPT-4o~\cite{Buker2025ChatbotsEndodontic} and similarly strong results for GPT-4, which also showed the lowest misinformation rate among compared models~\cite{Ozbay2025LLMEndodontics}. Accuracy increased further, to 95.0\% for ChatGPT-4o and Claude Opus 4, once the questions were drawn directly from AAE and ESE guideline statements rather than posed as open clinical queries~\cite{deAraujo2026EvidenceAIEndodontics}; even so, the same closed-format benchmark still produced confidently incorrect answers from its highest-scoring models, a hallucination risk the format's high accuracy alone does not capture. A similar pattern appears in trauma, where a guideline-based benchmark showed solid overall accuracy but a clear decline on more complex cases such as luxation injuries~\cite{Termteerapornpimol2025}, supporting the use of these tools as educational aids rather than diagnostic systems. The clearest version of this effect appears in standardized endodontic scenarios, where ChatGPT reached 99\% accuracy against dental students~\cite{Qutieshat2024AIEndodonticDiagnosis} — a result that says less about diagnostic ability than about how artificially clean a standardized scenario is compared to real clinical settings.

The third variable, which specific model or system is used, produces the largest differences of all. Eight chatbots evaluated on identical simulated periodontal cases of mucositis and peri-implantitis ranged from 88.8\% (GPT-4o) to 49.9\% (Copilot)~\cite{AmadorBarbosa2025}, a spread of nearly 39 percentage points on the exact same task, and ChatGPT-4.0 again led on open-ended periodontal questions~\cite{Chatzopoulos2025}. The same dependence on which model is chosen shows up in trauma, where ChatGPT-4o led on accuracy and DeepSeek led on completeness and readability, with no single model leading on every criterion at once~\cite{sezer}; Gemini, evaluated on a separate endodontic benchmark, showed significant improvement over a four-day testing period~\cite{Buker2025ChatbotsEndodontic}, meaning even a single model's standing is not always stable over time. This variable, unlike the first two, is also the one a developer can deliberately act on: PerioGPT, an instruction-tuned GPT-4o grounded in a RAG framework and a periodontal knowledge base, achieved high accuracy on specialist periodontal questions and AAP exam items while outperforming base chatbot models in accuracy, question complexity, and precision~\cite{Fanelli2025}, demonstrating that knowledge integration through RAG can substantially improve performance without retraining the underlying model. A lighter version of the same lever appears in implantology and oral and maxillofacial surgery, where advanced models already achieved high scores in question answering and case analysis~\cite{Wu2025LLMDentalImplantologyComparison}, and chain-of-thought prompting added a further, more modest 3.1\% gain on top of that baseline~\cite{Ji2025ChatGPTOralMaxillofacial} — refining performance that was already strong rather than fixing a weakness.

\subsubsection{Image-Based Diagnostic Interpretation}

Image-based evaluations present a more challenging picture than text-based reasoning, becoming less reliable as the visual precision required by the task increases. However, at the simplest level, where a well-defined structure is present in the image, customization and reasoning-focused processing produced clear improvements. A customized GPT-4V model reached 91\% accuracy for detecting supernumerary teeth, outperforming standard GPT-4o (77\%) and GPT-4V (63\%)~\cite{Asar2025}. Similarly, reasoning-focused multimodal models achieved 66--69\% sensitivity for implant fixture localization compared with 16.97\% for plain GPT-4o~\cite{Mine2026MLLMDentalImplants}.

Once the task requires localization rather than simple presence, reliability declines even where the improvements above held. Implant fixture localization, despite its higher sensitivity, still showed moderate accuracy, restoration-driven errors, and considerable run-to-run variability~\cite{Mine2026MLLMDentalImplants}. Periapical lesion detection, evaluated without comparable adaptation, showed the same difficulty more clearly: high sensitivity but very low specificity, indicating that the model could not reliably distinguish a true lesion from normal anatomy~\cite{Bubna2026LLMPeriapical}.

Reliability declines further once the task requires assigning a case to one of several visually similar categories. None of the evaluated chatbots reached acceptable agreement with expert classification of impacted mandibular third molars by Pell and Gregory grade~\cite{Erden2025ChatbotsImpactedMolars}, and labial frenulum attachment type classification from intraoral photographs showed similarly low, inconsistent accuracy~\cite{Kanmaz2025}. Diagnosis of mucosal lesions falls into this same category: overall diagnostic performance remained insufficient, although treatment recommendations were accurate once a correct diagnosis had already been reached~\cite{suarez2025diagnostic}. The clearest evidence that this is a limitation of visual representation rather than of prompting comes from oral lichen planus detection, where ordinary CNN models, built specifically for this kind of visual classification, outperformed every general-purpose LLM tested, even after prompting narrowed the gap~\cite{rewthamrongsris2025image}.

\subsubsection{Clinical Decision Support and Treatment Planning}
Language-generative models are increasingly being evaluated for clinical decision support in dentistry, with performance varying across tasks of different complexity.
At the administrative end, GPT-4-based systems reduced the time required for patient history analysis compared with conventional dentist-based evaluation~\cite{Kandaz2026}. This suggests potential value in workflow efficiency, particularly in tasks that do not require independent diagnostic or treatment decisions.

For safety-related decisions, findings were more mixed. In identifying drug--drug interactions, ChatGPT-5 achieved higher sensitivity but produced more false-positive alerts, whereas DeepSeek-Chat showed higher specificity but missed more critical interactions~\cite{Tayeb2025}. These findings highlight a clinically important trade-off between over-alerting and missed risk detection.

Guideline grounding appears to improve decision-support performance, although it does not eliminate model-dependent variation. In infective endocarditis prophylaxis, accuracy reached up to 90\% when guideline retrieval was integrated, and hallucinations were reduced~\cite{Rewthamrongsris2026}. In acute dental pain management, Claude and ChatGPT-4o produced the strongest performance, indicating that model selection remains important even when clinically relevant information is provided~\cite{Tosun2025}.

Treatment planning remains the most challenging area. In implantology, diagnostic accuracy exceeded treatment-planning accuracy~\cite{Wu2025LLMDentalImplantology}, while in trauma management, the best-performing model for diagnosis was not the best-performing model for antibiotic recommendation~\cite{Keles2026}. In endodontic restoration planning, none of the models achieved perfect repeatability across repeated assessments~\cite{Shirani2025}. Overall, these findings suggest that LLMs may support treatment planning, but clinician oversight remains essential.

\begin{table}[!htbp]
\centering
\scriptsize
\setlength{\tabcolsep}{4pt}
\caption{Summary of studies evaluating Clinical Reasoning and Diagnostic Support.}
\label{tab:llm_dental_1a}
\begin{adjustbox}{max width=\linewidth}
\begin{tabular}{
  >{\centering\arraybackslash}p{0.4cm}
  >{\RaggedRight}p{2.0cm}
  >{\RaggedRight}p{2.8cm}
  >{\RaggedRight}p{3.0cm}
  >{\RaggedRight}p{1.6cm}
  >{\RaggedRight}p{3.1cm}
  >{\RaggedRight}p{3.1cm}
}
\toprule
\textbf{Ref.} & \textbf{Domain} & \textbf{Objective} & \textbf{Models \& Methods} & \textbf{Dataset} & \textbf{Key Results} & \textbf{Limitations} \\
\midrule

\cite{Fanelli2025} & Periodontology & To develop and evaluate a (RAG) model for periodontal clinical applications & PerioGPT (GPT-4o with RAG) was trained using a large periodontal knowledge base and evaluated on clinical questions & 121 questions & PerioGPT achieved 81.16\% accuracy and generated more precise and complex responses compared with baseline LLMs & Requires domain-specific knowledge integration and curated datasets \\ \addlinespace[3pt]

\cite{tomo2024chatgpt} & Oral pathology & To evaluate the diagnostic accuracy and consistency of ChatGPT in generating differential diagnoses for oral and maxillofacial lesions based on clinical case descriptions & ChatGPT-3.5 and ChatGPT-4 were evaluated using structured text-based clinical scenarios, with regenerated prompts used to assess reproducibility and consistency of responses & 37 lesion cases & ChatGPT-4 achieved 80.18\% accuracy compared with 64.86\% for ChatGPT-3.5, approaching specialist performance (86.64\%) while demonstrating moderate consistency across repeated outputs & Output consistency remained moderate and strongly dependent on the completeness and quality of clinical descriptions \\ \addlinespace[3pt]

\cite{Buker2025ChatbotsEndodontic} & Endodontics & To assess accuracy, consistency, and temporal stability of chatbot responses to endodontic clinical questions & GPT-4o, Gemini, and Copilot were evaluated across multiple days and time points using validated question sets & 60 questions across different difficulty levels & GPT-4o achieved the highest overall accuracy (82.5\%) and showed more stable performance over time compared with other models & Responses varied across sessions and showed reduced reliability in complex scenarios \\ \addlinespace[3pt]

\cite{Ozbay2025LLMEndodontics} & Endodontics & To evaluate the accuracy and reliability of LLM-generated responses to clinically relevant endodontic questions & ChatGPT-3.5, ChatGPT-4, and Google Bard were assessed using expert Likert scoring of response quality & 40 open-ended questions & ChatGPT-4 provided the most accurate and informative responses and showed the lowest misinformation rate & All models produced incomplete or incorrect responses in certain clinical contexts \\ \addlinespace[3pt]

\cite{deAraujo2026EvidenceAIEndodontics} & Endodontics & To evaluate alignment of LLM outputs with evidence-based endodontic guidelines using standardized exam-style questions & Eleven LLMs were tested across multiple rounds using MCQs derived from AAE and ESE guidelines & 60 questions (3300 responses) & Top models such as ChatGPT-4o and Claude Opus achieved accuracy up to 95\%, demonstrating strong agreement with guideline-based answers & Performance varied between models and included confidently incorrect outputs \\ \addlinespace[3pt]

\cite{Termteerapornpimol2025} & Dental trauma & To benchmark the diagnostic knowledge and consistency of multiple LLMs across different traumatic dental injury categories & Seven LLMs including DeepSeek R1, ChatGPT-4o, and Gemini models were evaluated using repeated testing across five trauma subtopics with statistical analysis of accuracy and consistency & 125 questions with five repetitions per model & DeepSeek R1 achieved the highest overall accuracy (86.4\%), outperforming other models; however, performance decreased in complex categories such as luxation injuries and showed lower consistency compared to competitors & High variability across repetitions and reduced performance in complex clinical scenarios \\ \addlinespace[3pt]

\cite{Qutieshat2024AIEndodonticDiagnosis} & Endodontics & To compare diagnostic accuracy between AI models and dental students in endodontic case scenarios & ChatGPT-4 was compared with junior and senior dental students using standardized diagnostic cases & 7 clinical scenarios & ChatGPT achieved 99\% accuracy, significantly outperforming students who scored between 77–80\% & Potential overreliance on AI and reduced development of clinical reasoning skills \\ \addlinespace[3pt]

\cite{AmadorBarbosa2025} & Implantology & To evaluate diagnostic accuracy and treatment planning consistency of LLMs in peri-implant disease scenarios & Eight LLMs were tested using simulated clinical cases with repeated runs and expert scoring & 6 cases generating 720 outputs & GPT-4o achieved the highest diagnostic accuracy (88.8\%) and strongest consistency among models & Performance varied across runs and depended on prompt clarity \\ \addlinespace[3pt]

\cite{Chatzopoulos2025} & Periodontology & To assess the ability of LLMs to answer clinically relevant periodontal questions based on evidence-based criteria & GPT-4, Gemini, and Copilot were evaluated using expert scoring for accuracy and clinical relevance & 10 open-ended questions & ChatGPT-4 produced the most accurate and comprehensive responses, outperforming other models & Risk of inaccurate or misleading responses in complex clinical scenarios \\ \addlinespace[3pt]

\bottomrule
\end{tabular}
\end{adjustbox}
\end{table}

\begin{table}[!htbp]
\centering
\scriptsize
\setlength{\tabcolsep}{4pt}
\label{tab:llm_dental_1b}
\begin{adjustbox}{max width=\linewidth}
\begin{tabular}{
  >{\centering\arraybackslash}p{0.4cm}
  >{\RaggedRight}p{2.0cm}
  >{\RaggedRight}p{2.6cm}
  >{\RaggedRight}p{2.8cm}
  >{\RaggedRight}p{1.6cm}
  >{\RaggedRight}p{3.0cm}
  >{\RaggedRight}p{3.0cm}
}
\toprule
\textbf{Ref.} & \textbf{Domain} & \textbf{Objective} & \textbf{Models \& Methods} & \textbf{Dataset} & \textbf{Key Results} & \textbf{Limitations} \\
\midrule

\cite{sezer} & Pediatric dentistry & To evaluate and compare the performance of advanced LLMs in answering questions related to traumatic dental injuries in primary dentition based on international guidelines & ChatGPT-4o, DeepSeek, Gemini Advanced, and Claude 3.7 were assessed using open-ended trauma questions, with expert evaluation of accuracy, completeness, readability, and response time & 25 guideline-based questions & ChatGPT-4o demonstrated significantly higher accuracy than Gemini, while DeepSeek produced more complete and readable responses; strong correlation was observed between accuracy and completeness ($\rho = 0.701$) & Responses varied in readability and complexity, and none met ideal standards for clinical communication \\ \addlinespace[4pt]

\cite{Wu2025LLMDentalImplantologyComparison} & Implantology & To evaluate LLM performance in answering professional implantology questions and analyzing complex clinical scenarios requiring reasoning & GPT-based models, Gemini, DeepSeek, and Qwen were assessed using expert scoring for both structured questions and case-based reasoning tasks & 40 professional questions and 5 complex clinical cases & Gemini-2.0-flash-Thinking achieved the highest scores in both question answering and case analysis, indicating superior reasoning ability compared to other models & Performance varied across task types, with limitations in deep clinical reasoning and treatment planning consistency \\ \addlinespace[4pt]

\cite{Ji2025ChatGPTOralMaxillofacial} & OMFS & To evaluate whether chain-of-thought prompting improves LLM performance in oral and maxillofacial clinical questions & ChatGPT-4 was tested using both standard and CoT prompting across open-ended and multiple-choice questions, with expert evaluation of response quality & 130 open-ended and 1805 multiple-choice questions across multiple domains & CoT prompting improved overall performance by approximately 3.1\% and enhanced structure, completeness, and reasoning of responses & Despite improvements, performance remained below expert-level standards \\ \addlinespace[4pt]

\cite{Asar2025} & Radiology & To evaluate LLM performance in detecting supernumerary teeth in periapical radiographs & GPT-4V, GPT-4o, and a customized GPT-4V model trained using domain-specific references were compared & 180 radiographs & Customized GPT-4V achieved the highest diagnostic accuracy (91\%), outperforming standard GPT models & Performance depends on customization and availability of training data \\ \addlinespace[4pt]

\cite{Mine2026MLLMDentalImplants} & Radiology & To evaluate multimodal LLM performance in detecting and localizing dental implant fixtures in panoramic radiographs & GPT-4o, OpenAI o3, and GPT-5T were assessed using zero-shot image interpretation tasks, with evaluation based on sensitivity and specificity & 82 implant-present and 82 implant-absent panoramic radiographs & OpenAI o3 achieved the highest sensitivity (68.82\%), followed by GPT-5T (65.66\%), both outperforming GPT-4o; specificity varied across models, with improved performance in reasoning-based systems & Overall diagnostic performance remained moderate, with variability across runs and susceptibility to imaging artifacts \\ \addlinespace[4pt]

\cite{Bubna2026LLMPeriapical} & Endodontics & To evaluate LLM diagnostic performance in detecting periapical lesions on radiographic images & ChatGPT-5 and Gemini Flash were assessed using standardized multimodal prompts on periapical radiographs & 75 radiographs evenly distributed between lesion and non-lesion cases & ChatGPT-5 showed high sensitivity 97\% but very low specificity 0.11\%, indicating strong detection ability but high false-positive rates & Poor specificity and frequent false-positive classifications limit clinical applicability \\ \addlinespace[4pt]

\cite{Erden2025ChatbotsImpactedMolars} & OMFS & To evaluate chatbot performance in classifying impacted mandibular third molars using established radiographic classification systems & ChatGPT-4o, Gemini, Claude, and Copilot were assessed using panoramic radiographs and expert-defined classification systems including Winter and Pell–Gregory criteria & 93 impacted molar cases & ChatGPT-4o achieved the highest quality scores and showed better performance in certain classifications, but none of the models reached acceptable agreement with expert standards & Low agreement across classification systems and poor overall diagnostic reliability \\ \addlinespace[3pt]

\cite{Kanmaz2025} & Periodontology & To evaluate LLM accuracy in classifying labial frenulum attachment types from intraoral images & GPT-4o, Gemini, and Copilot were evaluated against expert consensus classifications & 117 images & All models demonstrated low accuracy, with maximum performance reaching 46.2\% & Poor agreement with expert standards and limited clinical utility \\ \addlinespace[4pt]

\bottomrule
\end{tabular}
\end{adjustbox}
\end{table}

\begin{table}[!htbp]
\centering
\scriptsize
\setlength{\tabcolsep}{4pt}
\label{tab:llm_dental_1c}
\begin{adjustbox}{max width=\linewidth}
\begin{tabular}{
  >{\centering\arraybackslash}p{0.4cm}
  >{\RaggedRight}p{1.5cm}
  >{\RaggedRight}p{2.8cm}
  >{\RaggedRight}p{3.0cm}
  >{\RaggedRight}p{1.6cm}
  >{\RaggedRight}p{3.1cm}
  >{\RaggedRight}p{3.1cm}
}
\toprule
\textbf{Ref.} & \textbf{Domain} & \textbf{Objective} & \textbf{Models \& Methods} & \textbf{Dataset} & \textbf{Key Results} & \textbf{Limitations} \\
\midrule

\cite{suarez2025diagnostic} & Oral pathology & To assess the diagnostic accuracy and repeatability of a multimodal LLM in identifying oral mucosal lesions and suggesting management & GPT-4o was evaluated using repeated image-based queries over time, with responses assessed by oral pathology experts & 30 clinical images (3600 responses) & Diagnostic accuracy was 58.2\%, with improved performance in lesion localization (71.4\%) and treatment recommendations (up to 95.8\%) when diagnosis was correct & Moderate diagnostic accuracy and risk of misclassification in complex cases \\ \addlinespace[4pt]

\cite{rewthamrongsris2025image} & Oral pathology & To evaluate multimodal LLM performance in detecting oral lichen planus and generating differential diagnoses from intraoral images & Seven LLMs were tested using zero-shot, example-guided, and differential prompting strategies and compared with CNN-based models & 1,142 histopathologically confirmed images & Example-guided prompting improved accuracy to 80.53\% with F1 score of 84.54\%, although CNN models consistently outperformed LLMs & LLMs showed inferior performance compared with specialized CNN models and were not suitable for independent diagnostic use \\ \addlinespace[4pt]

\cite{Kandaz2026} & General dentistry & To evaluate AI-assisted dental history analysis and decision support using structured prompting & GPT-4 API system (HistorAI) was applied to structured patient data & 77 patient cases with structured clinical data & AI significantly reduced analysis time compared with dentists while supporting decision-making & Requires further clinical validation \\ \addlinespace[4pt]

\cite{Tayeb2025} & Oral surgery & To compare LLMs and clinicians in detecting clinically relevant drug–drug interactions in dental therapy & ChatGPT-5, DeepSeek, and Gemini were evaluated against oral surgeons & 500 simulated clinical cases & GPT-5 showed high sensitivity (98\%), while DeepSeek showed perfect specificity (100\%) & Trade-off between sensitivity and specificity \\ \addlinespace[4pt]

\cite{Rewthamrongsris2026} & Cardio-dental & To evaluate RAG-based LLM performance in infective endocarditis prophylaxis decision-making & Multiple RAG-integrated LLMs were tested with and without prompting & 28 questions derived from AHA guidelines & Accuracy improved up to 90\% with optimized prompting strategies & Increased response time and variability \\ \addlinespace[4pt]

\cite{Tosun2025} & Endodontics & To evaluate LLM performance in managing acute dental pain based on clinical guidelines & GPT-4o, Claude, Gemini, and Copilot were evaluated using expert scoring & 20 open-ended questions based on ADA guidelines & GPT-4o and Claude demonstrated the highest performance in accuracy and completeness & Not sufficient to replace clinical guidelines \\ \addlinespace[4pt]

\cite{Wu2025LLMDentalImplantology} & Implantology & To assess the effectiveness of general LLMs in clinical consensus and implant case analysis using multiple evaluation metrics & ChatGPT-4, Gemini Pro 1.5, Claude, and Qwen were evaluated using Bayesian analysis and expert scoring across different types of clinical questions & Multiple clinical scenarios including simple and complex case analysis & ChatGPT-4 demonstrated the most stable performance across all tasks, while Gemini performed best in simple questions and Qwen showed high diagnostic scores but greater variability in treatment planning & Performance varied significantly depending on task complexity, particularly for treatment-related decisions \\ \addlinespace[4pt]

\cite{Keles2026} & Dental trauma & To evaluate diagnostic accuracy and treatment decision-making of AI chatbots in simulated dental trauma scenarios based on IADT guidelines & ChatGPT-4o, ChatGPT-5, DeepSeek, and Gemini were tested on standardized cases including clinical findings, radiographs, and pulp tests, with blinded expert scoring & 11 simulated trauma cases evaluated over 3 days & No statistically significant differences in overall diagnostic accuracy were found, although Gemini achieved 100\% accuracy; ChatGPT-4o showed highest accuracy in antibiotic indication, while ChatGPT-5 performed best in splinting decisions & High variability across different clinical parameters and inconsistency in treatment planning decisions \\ \addlinespace[4pt]

\cite{Shirani2025} & Restorative dentistry & To evaluate LLM performance in treatment planning for endodontically treated teeth over time & GPT-4.5, Gemini, Claude, and DeepSeek were evaluated longitudinally & 25 cases over 3 weeks & Gemini achieved the highest overall accuracy and consistency across sessions & Responses remained incomplete and inconsistent in some cases \\

\bottomrule
\end{tabular}
\end{adjustbox}
\end{table}

\subsection{Educational Applications}
Dental education is the most extensively studied application domain of generative AI in dentistry. Research has primarily focused on two areas: performance on licensing and specialty examinations (Table~\ref{tab:llm_dental_1d}) and use as an educational support tool for students and practitioners (Table~\ref{tab:llm_education}).

\subsubsection{Licensing and Specialty Examinations}

Large generative models have been evaluated across dental licensing, specialty, and dental technician examinations in diverse countries, and specialties. Across these contexts, a consistent pattern emerges: performance is strong on text-based questions but declines on image-based questions and on questions that require clinical reasoning.

This strength held across licensing and specialty exams in different countries and specialties, generally exceeding 80\% accuracy ~\cite{song2026kndle,Dashti2024,nguyen2025llm_dentistry,Dundar2026ChatGPTGeminiDentistry,Haberal2026AIDentists,Yilmaz2025AIOralPathology,Cakmak2025AIOrthodontic,Tassoker2025,wu2025oralpathology}. Nevertheless, performance remained below that of the highest-scoring human candidates where direct comparisons were available, particularly in clinically oriented domains~\cite{Sismanoglu2025}. Accuracy fell sharply wherever questions required clinical reasoning instead, most notably in exams such as Taiwan's national licensing examination and the more complex sections of specialty exams~\cite{lin2025taiwan,Akkoca2025,Haberal2026AIDentists,Cakmak2025AIOrthodontic}. Performance on dental technician examinations was also moderate, with only limited evidence of self-learning improvement in some models~\cite{huang2025tndtle,fukuda2025jndte}.

A separate limitation emerges once a question depends on interpreting an image rather than recalling knowledge, as shown in the Japanese National Dental Examination, where text-based accuracy reached 79.9–92.2\%, while accuracy on the image-based questions fell to 45.6–67.8\%~\cite{mine2025benchmarking}, a gap also reported in image-dependent oral pathology questions~\cite{watanabe2025oralpathology}. This indicates the limitation lies in visual interpretation rather than dental knowledge. Together, these findings suggest strong scores on text-based, recall-oriented questions may not translate into the image interpretation and clinical decision-making skills real practice demands.

\subsubsection{Language-Generative Models as Dental Education Support Tools}

Large generative models have been evaluated in dental education across distinct roles such as knowledge assessment, learning support, and educational content generation, with performance profiles that vary accordingly.

As knowledge and assessment tools, they perform strongly on undergraduate-level and diagnostic case-based questions in dental occlusion and endodontics, in some cases exceeding 90\% accuracy and outperforming dental students~\cite{Alqahtani2025AI, AriliOzturk2025ChatGPT, Durmazpinar2025Diagnostic}. Beyond these domains, performance shows a more moderate profile. It declines with content complexity and varies across question formats, though it remains sufficiently consistent to support guided learning rather than independent clinical decision-making~\cite{Azhari2026AI, Kurt2025Knowledge, Saglam2026AI, Kuru2025AI, LlorenteDePedro2025ChatGPT}. This suggests that question difficulty and format are stronger drivers of performance than the dental topic alone.

Content generation appears more robust. AI-generated clinical case simulations produced outcomes comparable to real patient interactions with greater standardization, information density, and transparency of reasoning~\cite{RodriguesPereira2025AI}, and LLMs generated reflective assignments largely comparable to human work~\cite{Brondani2024AI}. Unlike question-answering, where reliability varies with complexity, content generation is notably more consistent---suggesting these models are better suited to producing structured educational material than to answering complex clinical questions.

Performance varies across dental specialties, with generally accurate responses but limitations in reliability, consistency across repeated testing, and distinguishing closely related concepts~\cite{Dermata2025AI, Hakami2026AI, Koyuncuoglu2025AI, Lafourcade2025LLM}, though strategies such as RAG, ICL, and majority voting have been shown to improve performance and reduce knowledge-based errors~\cite{Gao2026LLM}. Overall, LLMs provide accessible, high-quality informational content for students and professionals, although challenges related to readability and reliability remain~\cite{Ozturk2025ChatGPT}.

\begin{table}[!htbp]
\centering
\scriptsize
\setlength{\tabcolsep}{4pt}
\caption{Summary of studies evaluating LLMs in Licensing and Specialty Examinations.}
\label{tab:llm_dental_1d}
\begin{adjustbox}{max width=\linewidth}
\begin{tabular}{
  >{\centering\arraybackslash}p{0.4cm}
  >{\RaggedRight}p{1.5cm}
  >{\RaggedRight}p{2.8cm}
  >{\RaggedRight}p{3.0cm}
  >{\RaggedRight}p{1.6cm}
  >{\RaggedRight}p{3.1cm}
  >{\RaggedRight}p{3.1cm}
}
\toprule
\textbf{Ref.} & \textbf{Domain} & \textbf{Objective} & \textbf{Dataset} & \textbf{Models} & \textbf{Key Results} & \textbf{Limitations} \\
\midrule
\cite{song2026kndle} & Korean National Dental Licensing Examination (KNDLE) & To compare accuracy and consistency of GPT-4o and Gemini on licensing exam questions & 1,401 text-based MCQs from KNDLE (2019–2023), answered in three runs & GPT-4o, Gemini Advanced & GPT-4o: 81.1\%, Gemini: 76.6\% (significant difference) & Both models performed better in basic sciences than in clinical subjects \\ \addlinespace[3pt]

\cite{Dashti2024} & USA – INBDE, ADAT, DAT & To evaluate ChatGPT performance on multiple U.S. dental examinations & 253 questions including knowledge-based and case history questions from INBDE, ADAT, and DAT & ChatGPT-3.5, ChatGPT-4 & INBDE: $\sim$80\%; ADAT: 66–83\%; DAT: up to 94\% & Higher accuracy in knowledge-based and comprehension questions \\ \addlinespace[3pt]

\cite{nguyen2025llm_dentistry} & USA-based dental MCQs & To compare accuracy of multiple LLMs on dental MCQs including image-based questions & 1,490 MCQs from U.S. National Board Dental Examination review books & ChatGPT, Claude, Copilot, Gemini, Mistral, Llama & Highest accuracy models $\sim$83–85\% & Significant differences between models; lower accuracy in image-based questions \\ \addlinespace[3pt]

\cite{Dundar2026ChatGPTGeminiDentistry} & Turkey – Dental Specialty Examination (DUS) & To compare performance of ChatGPT-4o and Gemini Advanced across multiple disciplines & 1,504 MCQs from DUS (2012–2021), including basic and clinical sciences & ChatGPT-4o, Gemini Advanced & ChatGPT-4o: 84\%, Gemini: 81.8\% & No significant overall difference; variation across disciplines \\ \addlinespace[3pt]

\cite{Haberal2026AIDentists} & Turkey – DUS (Restorative dentistry) & To evaluate and compare multiple AI chatbots on restorative dentistry exam questions & 188 MCQs from DUS exams (2012–2025) & ChatGPT variants, Gemini, Claude, Copilot, DeepSeek & Highest accuracy: Gemini Advanced (96.28\%), ChatGPT-4o Plus (93.62\%) & High overall accuracy with differences across models \\ \addlinespace[3pt]

\cite{Yilmaz2025AIOralPathology} & Turkey – DUS (Oral pathology) & To compare performance of multiple LLMs on oral pathology MCQs & 100 oral pathology MCQs from DUS (2012–2021) & ChatGPT-4o, Gemini, Claude, DeepSeek, ChatGPT-o1 & Highest accuracy: ChatGPT-o1 (96\%) & Significant differences between models; variation by question type \\ \addlinespace[3pt]

\cite{Cakmak2025AIOrthodontic} & Turkey – DUS (Orthodontics) & To evaluate chatbot performance and reliability of sources in orthodontic questions & 129 orthodontic MCQs categorized by cognitive level (Bloom's taxonomy) & ChatGPT-5, Claude 3.7, Copilot & Accuracy: 82.2–85.3\% & Higher performance in scenario-based questions; lower in visual analysis \\ \addlinespace[3pt]

\cite{Tassoker2025} & Turkey – DUS (Oral radiology) & To compare chatbot performance in oral radiology MCQs & 123 text-based oral radiology MCQs from DUS (2012–2021); image questions not included & ChatGPT-4o, Bard, Copilot & ChatGPT-4o: 86.1\%, Bard: 61.8\%, Copilot: 41.5\% & Significant differences in accuracy between models \\\addlinespace[3pt]

\cite{wu2025oralpathology} & Taiwan – Oral pathology (NDLE) & To evaluate ChatGPT performance on oral pathology MCQs of different types & 280 oral pathology MCQs from Taiwan licensing exam & ChatGPT-4, ChatGPT-4o (with and without prompt) & Highest accuracy: 90\% (ChatGPT-4o with prompt) & Performance varied across question types; prompting improved accuracy \\ \addlinespace[3pt]

\cite{Sismanoglu2025} & Turkey – DUS vs human candidates & To compare performance of AI chatbots with human exam performance & DUS exams (2020–2021), including basic and clinical sections & ChatGPT-4, Gemini Advanced & Both models exceeded passing threshold but scored lower than top candidates & AI performance lower than highest-performing human candidates \\ \addlinespace[3pt]

\cite{lin2025taiwan} & Taiwan – National Dental Licensing Examination (NDLE) & To assess performance of AI chatbots on licensing exam questions & 2,699 MCQs across basic and clinical dentistry subjects; image questions excluded & ChatGPT-3.5, Gemini, Claude2 & Claude: 54.89\%, ChatGPT: 49.33\%, Gemini: 44.63\% & None of the models achieved passing scores \\ \addlinespace[3pt]

\cite{Akkoca2025} & Oral radiology (education, textbook-based) & To assess ChatGPT-4o performance in answering radiology questions over time & 99 questions from oral radiology textbook answered repeatedly over 10 days & ChatGPT-4o & Overall accuracy: 59.4\% & Performance varied depending on time of testing \\

\bottomrule
\end{tabular}
\end{adjustbox}
\end{table}
\begin{table}[!htbp]
\centering
\scriptsize
\setlength{\tabcolsep}{4pt}
\label{tab:llm_dental_1e_and_education}

\begin{adjustbox}{max width=\linewidth}
\begin{tabular}{
  >{\centering\arraybackslash}p{0.4cm}
  >{\RaggedRight}p{1.5cm}
  >{\RaggedRight}p{2.8cm}
  >{\RaggedRight}p{3.0cm}
  >{\RaggedRight}p{1.6cm}
  >{\RaggedRight}p{3.1cm}
  >{\RaggedRight}p{3.1cm}
}
\toprule
\textbf{Ref.} & \textbf{Domain} & \textbf{Objective} & \textbf{Dataset} & \textbf{Models} & \textbf{Key Results} & \textbf{Limitations} \\
\midrule

\cite{huang2025tndtle} & Taiwan – Dental technician exam & To evaluate accuracy and improvement over time (self-learning ability) of LLMs & 194 text-based MCQs from 2023 technician licensing exam, repeated over three weeks & ChatGPT-4, Gemini, DeepSeek-V3 & Initial accuracy: 52.1\%–69.6\%; Gemini showed improvement over time & Gemini demonstrated performance improvement, others showed limited change \\ \addlinespace[4pt]
\cite{fukuda2025jndte} & Japan – Dental technician examination & To evaluate ChatGPT-4o accuracy on technician licensing exam questions & 400 questions from Japanese National Dental Technician Examination (2018–2022) & ChatGPT-4o & Highest accuracy: 84.1\%; lowest: 46.2\% depending on subject & Higher accuracy in basic knowledge domains and lower in specialized technical fields \\ \addlinespace[4pt]

\cite{mine2025benchmarking} & Japan – Japanese National Dental Examination (JNDE 2024) & To evaluate performance of multimodal LLMs on licensing exam questions including text and visual items & 353 questions from JNDE-2024, including 204 text-only and 149 visually based questions across 17 dental specialties & ChatGPT-4o, OpenAI o1, Claude 3.5 Sonnet, Gemini & Overall accuracy: o1 (81.9\%), Claude (71.7\%), Gemini (66.6\%), GPT-4o (65.7\%); text questions: 79.9–92.2\%, visual questions: 45.6–67.8\% & Models performed significantly better on text-only questions than visually based questions \\ \addlinespace[4pt]

\cite{watanabe2025oralpathology} & Japan – Oral pathology (JNDE images) & To evaluate diagnostic accuracy of LLMs on oral pathology image-based questions & 176 pathology image-based questions from Japanese National Dental Examination  & ChatGPT-4o, Gemini 1.5 Pro, Claude 3.5 Sonnet & Accuracy: Gemini (61.4\%), Claude (52.3\%), ChatGPT-4o (45.4\%) & LLMs showed moderate diagnostic performance with variation across disease categories \\ \addlinespace[4pt]

\bottomrule
\end{tabular}
\end{adjustbox}



\vspace{1.5em}
\captionof{table}{Summary of studies evaluating LLMs as Dental Education Support Tools.}
\label{tab:llm_education}
\begin{adjustbox}{max width=\linewidth}
\begin{tabular}{
  >{\centering\arraybackslash}p{0.4cm}
  >{\RaggedRight}p{2.8cm}
  >{\RaggedRight}p{3.0cm}
  >{\RaggedRight}p{1.7cm}
  >{\RaggedRight}p{3.0cm}
  >{\RaggedRight}p{3.0cm}
}
\toprule
\textbf{Ref.} & \textbf{Objective} & \textbf{Models \& Methods} & \textbf{Dataset} & \textbf{Key Results} & \textbf{Key Findings} \\
\midrule

\cite{Alqahtani2025AI} & To evaluate chatbot performance on dental occlusion MCQs and assess consistency & Claude 3.7, GPT-4o, DeepSeek, and LLaMA tested across two rounds; agreement assessed with Kappa and McNemar tests & Standardized MCQs from dental occlusion textbook & Claude 3.7 and GPT-4o achieved highest accuracy (90–91.4\%); LLaMA lowest (64.3\%) & Claude and GPT-4o showed strong reliability; performance varied across models \\ \addlinespace[4pt]

\cite{AriliOzturk2025ChatGPT} & To evaluate accuracy and consistency of ChatGPT-4 and ChatGPT-4o in undergraduate endodontic education & Questions asked multiple times daily over three days; statistical analysis of accuracy and consistency & 60 MCQs from six undergraduate endodontic topics & ChatGPT-4o achieved higher accuracy (92.8\%) than ChatGPT-4 (81.7\%) (p<0.001) & ChatGPT-4o showed improved accuracy, supporting use as educational tool with limitations \\ \addlinespace[4pt]

\cite{Durmazpinar2025Diagnostic} & To compare diagnostic accuracy of ChatGPT-4o with dental students in endodontic cases & ChatGPT-4o and student groups evaluated using radiographs, clinical photos, and histories & 15 case-based MCQs based on AAE guidelines & ChatGPT-4o achieved 91.4\% accuracy, outperforming both student groups (p<0.001) & AI demonstrated strong diagnostic support potential in education \\ \addlinespace[4pt]

\cite{Azhari2026AI} & To compare the accuracy of ChatGPT and Google Gemini in answering dental caries MCQs using simulated examinations & ChatGPT-3.5 and Gemini evaluated using standardized prompts across seven simulated exam lengths with statistical analysis (t-test, ANOVA) & 125 validated dental caries MCQs; 7 exam groups (25–85 questions); 1400 simulations & Gemini significantly outperformed ChatGPT (p<0.001); higher mean scores and passing rates across all exam formats & Gemini demonstrated more stable and reliable performance, although both models struggled with complex content \\ \addlinespace[4pt]
 
\bottomrule
\end{tabular}
\end{adjustbox}
\end{table}

\begin{table}[!htbp]
\centering
\scriptsize
\setlength{\tabcolsep}{4pt}
\label{tab:llm_education_2}
\begin{adjustbox}{max width=\linewidth}
\begin{tabular}{
  >{\centering\arraybackslash}p{0.4cm}
  >{\RaggedRight}p{2.8cm}
  >{\RaggedRight}p{3.2cm}
  >{\RaggedRight}p{2.2cm}
  >{\RaggedRight}p{3.6cm}
  >{\RaggedRight}p{3.1cm}
}
\toprule
\textbf{Ref.} & \textbf{Objective} & \textbf{Models \& Methods} & \textbf{Dataset} & \textbf{Key Results} & \textbf{Key Findings} \\
\midrule

\cite{Kurt2025Knowledge} & To compare AI chatbots and dental students in diagnosing pulpal and periapical diseases & ChatGPT-3.5, ChatGPT-4o, Gemini, Copilot compared with students using statistical tests & 15 MCQs based on AAE guidelines & ChatGPT-4o (79.6\%) performed close to senior students (85.1\%) & ChatGPT-4o demonstrated comparable performance to experienced students \\ \addlinespace[4pt]

\cite{Saglam2026AI} & To compare knowledge levels of dentists and AI chatbots in traumatic dental injury management & ChatGPT-4o and Gemini compared with dentists using MCQ-based questionnaire & 20 MCQs based on AAE trauma guidelines & ChatGPT-4o provided significantly more correct answers ($p < 0.05$); highest knowledge level (100\%) & AI outperformed dentists but still not fully sufficient for clinical use \\ \addlinespace[4pt]

\cite{Kuru2025AI} & To evaluate reliability of multiple LLMs in answering dental trauma questions & Five LLMs evaluated over nine days using repeated questioning and statistical tests & 30 trauma questions (MCQ, fill-in, dichotomous; 1350 responses) & No significant difference between models ($p>0.05$); ChatGPT-3.5 highest accuracy (76.7\%) & LLMs show potential but require cautious use due to variability \\ \addlinespace[4pt]

\cite{LlorenteDePedro2025ChatGPT} & To assess reliability and repeatability of ChatGPT responses in endodontics & ChatGPT-4 and 4o tested repeatedly (30 times per question); evaluated by experts & 30 guideline-based clinical questions (1800 responses) & ChatGPT-4o showed slightly higher reliability (55.22\%) than ChatGPT-4 (52.67\%) & Both models showed moderate reliability; supervised use recommended \\ \addlinespace[4pt]

\cite{RodriguesPereira2025AI} & To compare AI-driven clinical simulation with real patient-based learning in TMD education & ChatGPT-3.5 simulation compared with real patient cases in crossover study & 30 students completing two simulation scenarios & AI produced higher information density and similar diagnostic accuracy ($p>0.05$) & AI simulations matched real patients in diagnostic reasoning and improved clarity \\ \addlinespace[4pt]

\cite{Brondani2024AI} & To evaluate ChatGPT-generated reflections and qualitative analysis in dental education & ChatGPT-generated and student-generated reflections compared by instructors and researchers & 20 reflections; subset analyzed qualitatively & Instructors correctly identified authorship 85\% of the time; thematic analysis similar to human researchers & AI can produce comparable reflective and qualitative outputs in education \\ \addlinespace[4pt]

\cite{Dermata2025AI} & To evaluate diagnostic accuracy and educational potential of LLMs in paediatric dentistry & Six LLMs evaluated using rubric scoring and statistical comparison & 10 open-ended clinical questions & ChatGPT-4 achieved highest score (8.08), followed by Gemini Advanced (8.06) & LLMs show educational potential but require cautious interpretation \\ \addlinespace[4pt]

\cite{Hakami2026AI} & To assess chatbot performance in orthodontic knowledge questions & Five chatbots evaluated across two rounds and compared with students & 80 orthodontic MCQs & DeepSeek performed best; Meta AI lowest; improvements observed in some models in second round & Chatbot performance varied and required further refinement \\ \addlinespace[4pt]

\cite{Koyuncuoglu2025AI} & To evaluate reliability of chatbot responses in periodontology exam questions over time & Chatbots tested twice with one-month interval using statistical comparison & 125 exam questions across multiple formats & CoPilot Pro highest accuracy (73.6--75.2\%); GPT-4o showed inconsistency & AI responses varied over time; reliability concerns remain \\ \addlinespace[4pt]

\cite{Lafourcade2025LLM} & To evaluate accuracy, consistency, and contextual understanding of LLMs in restorative dentistry and endodontics & Multiple LLMs evaluated with and without contextual input & 517 MCQs and 539 definitions from educational archives & ChatGPT-4 and Claude-3 achieved higher accuracy and repeatability; contextual input improved performance & LLMs showed moderate performance with variability across concepts \\ \addlinespace[4pt]

\cite{Gao2026LLM} & To evaluate enhanced LLMs using RAG, ICL, and majority voting on prosthodontic MCQs & Enhanced and base models compared using statistical analysis & Standardized prosthodontic MCQs (Chinese and English) & Enhanced models significantly improved accuracy in Chinese MCQs (p<0.001) & Model enhancement strategies improved performance but inconsistently across languages \\ \addlinespace[4pt]

\cite{Ozturk2025ChatGPT} & To evaluate quality and reliability of ChatGPT responses in traumatic dental injuries for students and professionals & ChatGPT-3.5 and 4 evaluated using multiple scoring systems (GQS, DISCERN, readability indices) & 40 trauma-related questions & Both models produced high-quality information but were difficult to read; ChatGPT-4 higher quality in diagnosis section & LLMs provide useful information but require additional sources for reliability \\

\bottomrule
\end{tabular}
\end{adjustbox}
\end{table}

\subsection{Patient Communication and Report Generation}
The third major use of language-generative models in dentistry is patient 
communication, spanning two areas: patient question answering and radiology 
report generation, with findings summarized in 
Tables~\ref{tab:llm_patient_question_answering} and~\ref{tab:llm_report_generation}. 
This includes answering questions, creating consent forms, simplifying reports, 
and helping patients make decisions. It is a sensitive area because clear, 
accurate, and empathetic information affects safety and how well patients 
follow treatment.

\subsubsection{Patient Question Answering}
Patients increasingly turn to online resources for dental information and self-management. In this context, LLMs demonstrate consistent advantages over traditional search engines in response quality, readability, empathy, and patient satisfaction~\cite{Ren2026LLMOrthodontic}, with newer models generally outperforming earlier systems in accuracy and comprehensiveness across dental specialties~\cite{Salmanpour2025, Dursun2024, Erdem2025}. Performance declines as question complexity increases, and readability remains inconsistent across models and contexts~\cite{zhang2025comprehensiveness, sezer2025chatgpt_pediatric, Yamac2025, kofos2025evaluation}.

An analytically important finding concerns how differently clinicians and patients evaluate the same responses. Studies in postoperative and follow-up settings report high ratings for accuracy, appropriateness, and empathy~\cite{cai2024chatgpt_followup, batool2024llm_dental_care}, yet when both groups assess the same outputs, clinicians consistently prefer expert-generated content while patients and parents sometimes rate chatbot responses as comparable or superior~\cite{celik2026bridging}. This divergence suggests LLMs may satisfy patient communication expectations — clarity, accessibility, empathy — without necessarily meeting clinical standards for precision and completeness.

Errors, inconsistencies, and hallucinated information remain recurring risks across evaluations~\cite{Chen2025, Prasad2025}, positioning LLMs as supplementary communication tools rather than independent sources of dental advice.

\subsubsection{Radiology Report Generation}

LLMs can substantially reduce the time required for radiology report writing and produce readable, well-structured output~\cite{Stephan2024}. However, fluent language does not reflect diagnostic completeness, and critical details may be absent even in well-written reports. This limitation can be addressed by integrating vision-based models with LLMs to separate image interpretation from text generation, an approach that improves report quality and reduces hallucinations~\cite{Dasanayaka2025, Balel2026}.

Simplification of radiology reports for patients represents a more consistently successful application, improving readability, understanding, and engagement, facilitating patient participation in clinical discussions and decision-making~\cite{Stephan2025}, with LLM-generated communications generally rated as more structured, clear, and empathetic than conventional alternatives~\cite{Maruska2025, Ren2026LLMOrthodontic}. Readability inconsistency and insufficient actionable guidance remain limitations in some contexts~\cite{Alnsour2025}, indicating that even this more promising application requires additional review before routine deployment.

\begin{figure}[!htbp]
    \centering
\includegraphics[width=\linewidth]{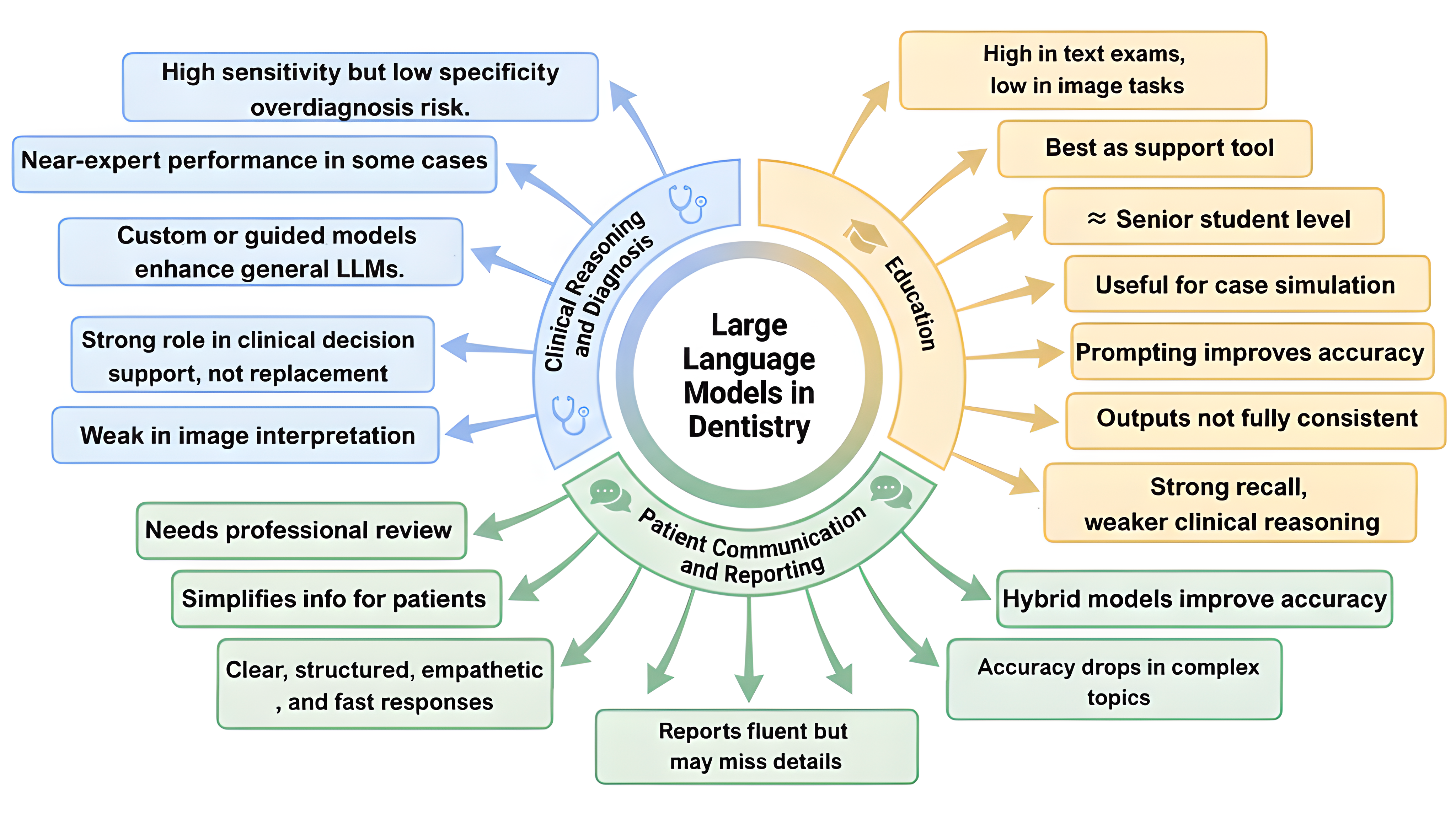}
    \caption{Summary of the main themes across clinical, educational, and patient communication domains.}
    \label{fig:llm_summary}
\end{figure}

Figure~\ref{fig:llm_summary} summarizes the main points across clinical, educational, and communication aspects of language-generative models in dentistry.

\begin{table}[!htbp]
\centering
\tiny
\setlength{\tabcolsep}{2pt}
\setcounter{table}{3}
\caption{Summary of studies evaluating LLMs in Patient Question Answering.}
\label{tab:llm_patient_question_answering}
\begin{adjustbox}{max width=\linewidth}
\begin{tabular}{
  >{\centering\arraybackslash}p{0.5cm}
  >{\RaggedRight}p{2.4cm}
  >{\RaggedRight}p{2.2cm}
  >{\RaggedRight}p{2.4cm}
  >{\RaggedRight}p{2.6cm}
  >{\RaggedRight}p{2.7cm}
  >{\RaggedRight}p{2.7cm}
}
\toprule
\textbf{Ref.} & \textbf{Objective} & \textbf{Dataset} & \textbf{Methods} & \textbf{Key Results} & \textbf{Conclusion} & \textbf{Limitations} \\
\midrule

\cite{Ren2026LLMOrthodontic} &
Compare 5 LLMs vs.\ 3 search engines for orthodontic patient questions &
45 questions (6 categories) &
Expert Likert scoring (quality, empathy, readability, satisfaction) &
LLMs $>$ search engines: quality (4.00 vs.\ 3.50), satisfaction (8.00 vs.\ 7.25); GPT-4o best &
LLMs are effective supplementary tools for patient education and self-management &
Single time-point; ICC\,=\,0.68 \\ \addlinespace[4pt]

\cite{Salmanpour2025} &
Compare ChatGPT-4.0 and Copilot vs.\ expert responses to orthodontic FAQs &
15 AAO FAQs; 52 orthodontists + 102 patients &
Ratings by orthodontists and patients (10-point adequacy scale) &
Experts highest; both chatbots acceptable ($\geq$6/10); Copilot lowest; both below experts ($p{<}0.05$) &
Chatbots useful but do not yet match expert adequacy &
Limited generalizability; two chatbot models only; lacks personalization \\ \addlinespace[4pt]

\cite{Dursun2024} &
Assess AI chatbots  (ChatGPT-3.5, ChatGPT-4, Gemini, Copilot) as patient information tools for clear aligners &
20 frequently asked patient questions on clear aligners (from 180 pooled across 30 websites) &
Likert; modified DISCERN; GQS; FRES &
ChatGPT-4 best accuracy (4.5); Copilot best DISCERN/GQS; Gemini best readability (FRES\,=\,54.12); others FRES 38--44 &
AI chatbots useful; need improved readability &
English-only; single time-point; limited generalizability\\ \addlinespace[4pt]

\cite{Erdem2025} &
Evaluate the performance of chatbots (ChatGPT-3.5, ChatGPT-4.0, Copilot, Gemini) in orthodontic emergency scenarios &
40 orthodontic emergency questions (7 categories)  &
5-point Likert; 3 orthodontic experts &
Gemini and ChatGPT-4.0 highest; ChatGPT-3.5 lowest; only Fixed Orthodontic Treatment significant ($p{=}0.043$) &
Chatbots can assist in emergencies; accuracy varies by model &
Single time-point evaluation; responses may vary with real-time data changes \\ \addlinespace[4pt]

\cite{zhang2025comprehensiveness} &
Assess ChatGPT-3.5 and ChatGPT-4 comprehensiveness on gingival and endodontic queries &
33 questions (17 common-sense, 16 expert); English \& Chinese &
5-point Likert; 3 specialists &
GPT-4 (4.45) $>$ GPT-3.5 (4.03); English (4.53) $>$ Chinese (3.95); common-sense (4.46) $>$ expert-level (4.02); all $p{<}0.05$&
GPT-4 more comprehensive; both better in English and common-sense queries &
Limited question set; small expert panel; findings time-sensitive due to LLM evolution\\ \addlinespace[4pt]

\cite{sezer2025chatgpt_pediatric} &
Evaluate ChatGPT-4 responses to pediatric dentistry questions &
30 FAQs + 30 curricular questions; 30 pediatric dentists &
5-point Likert (accuracy); 3-point (completeness) &
FAQ accuracy 4.21\,$\pm$\,0.55 vs.\ curricular 4.16\,$\pm$\,0.70 (ns, $p{=}0.942$); topic variation ($p{=}0.007$): fissure sealants highest (4.45\,$\pm$\,0.62), pulpal therapy lowest (3.93\,$\pm$\,0.93) &
ChatGPT-4 performs well on standardized topics; less reliable on complex areas (fluoride, pulpal therapy); expert oversight essential &
No evaluator calibration; inter-rater reliability not formally calculated; FAQs not from actual patients \\ \addlinespace[4pt]

\cite{Yamac2025} &
Evaluate four LLMs( ChatGPT-3.5, ChatGPT-4, Google Bard, Bing Chat)  responses on orthodontic miniscrews &
30 patient questions on orthodontic miniscrews sourced from Google search &
5-point Likert + mDISCERN; 3 orthodontic residents &
GPT-4 highest Likert (3.84); GPT-4/Bard comparable mDISCERN (${\approx}$23.4); Bing Chat lowest (3.37 / 21.63) &
LLMs can inform patients on miniscrews; professional supplementation necessary &
No blinding of chatbots before evaluation; no assessment of response changes over time \\ \addlinespace[4pt]

\cite{kofos2025evaluation} &
Compare responses of ChatGPT (GPT-4), Gemini, and MedGebra to dental implant patient questions&
56 EAO patient-guide questions on dental implants &
4-point Likert (accuracy, precision, clarity); 2 oral surgery specialists &
ChatGPT highest, particularly in clarity; Gemini lower accuracy/precision; MedGebra lowest on all criteria &
LLMs support implant patient education; professional oversight crucial & English-only; no evaluator calibration ($\kappa$\,=\,0.457); structured questions may not reflect real patient interactions \\ \addlinespace[4pt]

\cite{cai2024chatgpt_followup} &
Assess ChatGPT/GPT-4 for post-surgical oral surgery follow-up &
30 commonly asked post-operative follow-up questions &
Evaluation by 3 oral and maxillofacial surgeons; accuracy and rationality scoring (0--10 scale)&
Full marks for accuracy and rationality; sensed patient emotions and provided reassurance &
LLMs effective for post-surgical follow-up under professional guidance &
Single-round conversations only; vague time-frame responses may not satisfy patients\\ \addlinespace[4pt]

\cite{batool2024llm_dental_care} &
Compare embedded GPT vs ChatGPT-3.5 in post-operative dental care &
40 questions across 4 dental specialties  &
4-point Likert scale; CVI ; modified kappa ($K^*$); 36 dental experts (9 per specialty) &
Embedded GPT: CVI 65.62\%, accuracy 62.5\%, clarity 72.5\%; ChatGPT-3.5: CVI 61.87\%, accuracy 52.5\%, clarity 67.5\% &
Embedding and prompt engineering improve post-operative dental care responses; further optimization needed &
Limited to ChatGPT only; questions not from actual patients; subjective specialist evaluation \\ \addlinespace[4pt]

\cite{celik2026bridging} &
Compare AI (ChatGPT-4o, Gemini Advanced) vs expert responses in pediatric dentistry; parent vs dentist perception &
15 IAPD FAQs (screened from 57 by 20 dentists) &
Flesch-Kincaid readability; 1--10 adequacy scale; blinded survey by 47 pediatric dentists + 101 parents &
Dentists rated experts higher in 13/15 ($p{<}0.05$); parents showed no significant difference in 8/15; AI rated comparably or higher than experts by parents in some cases &
AI--expert alignment inconsistent; chatbots show promise for patient education; expert oversight remains essential &
only two chatbot models were tested; no prompt optimization applied \\ \addlinespace[4pt]

\cite{Chen2025} &
Evaluate LLMs (Ernie Bot, ChatGPT, Gemini) responses to orthodontic pre-treatment consultation questions &
50 pre-treatment consultation questions from 1,762 real clinical queries &
6 dimensions (professionalism, content accuracy, clarity, personalization, completeness, empathy); 5-point scale; 4 experts + 4 patients&
All models scored 3.52--4.08/5; Ernie Bot highest in IC (4.08) and PR (3.98)&
LLMs show capability in open-ended questions but require caution in orthodontic consultation &
Semantic translation errors; cultural bias; temporal context unspecified\\ \addlinespace[6pt]

\cite{Prasad2025} &
Assess readability and accuracy of AI chatbots( ChatGPT-3.5, Copilot, Gemini) vs.\ ACP content for prosthodontic patient education &
26 ACP FAQs (dentures, implants, oral cancer, restorations, prosthodontist, brushing/flossing) &
FRES; FKGL; Likert-based accuracy, completeness, quality &
ChatGPT highest word count, ACP lowest; prosthodontist least readable; brushing/flossing most readable and accurate; denture accuracy lowest ($p{=}0.006$)&
LLMs support prosthodontic education; professionals must critically evaluate AI content &
Unblinded evaluator panel; limited prosthodontic topics; AI performance may change over time \\

\bottomrule
\end{tabular}
\end{adjustbox}
\end{table}

\begin{table}[!htbp]
\centering
\scriptsize
\setlength{\tabcolsep}{4pt}
\caption{Summary of studies evaluating LLMs in Report Generation.}
\label{tab:llm_report_generation}
\begin{adjustbox}{max width=\linewidth}
\begin{tabular}{
  >{\centering\arraybackslash}p{0.5cm}
  >{\RaggedRight}p{2.4cm}
  >{\RaggedRight}p{2.2cm}
  >{\RaggedRight}p{2.3cm}
  >{\RaggedRight}p{2.7cm}
  >{\RaggedRight}p{2.7cm}
  >{\RaggedRight}p{2.7cm}
}
\toprule
\textbf{Ref.} & \textbf{Objective} & \textbf{Dataset} & \textbf{Methods} & \textbf{Key Results} & \textbf{Conclusion} & \textbf{Limitations} \\
\midrule

\cite{Stephan2024} &
Evaluate ChatGPT for radiology report generation from panoramic radiographs &
100 dental students generating reports from panoramic radiographs &
Comparison between manually written reports and ChatGPT-generated reports using standardized prompts &
AI-generated reports showed high textual similarity to reference reports and were largely error-free, but lacked critical diagnostic information compared to student-written reports &
ChatGPT demonstrates potential for automated radiology report generation &
difficulty interpreting complex odontogram data \\ \addlinespace[6pt]

\cite{Dasanayaka2025} &
Develop a multimodal system( (BLIP-2 + Llama 3 8B) for orthopantomography radiology report generation and Q\&A &
4 custom datasets: DPT images (1000 pairs), radiology reports, Sinhala corpus, Sinhala dental Q\&A &
BLIP-2-based captioning combined with Llama 3 LLM and expert evaluation &
Overall accuracy 81.3\%; caries 87.9\%, impacted teeth 89.7\%, bone loss 88\%, periapical lesions 81.8\%; subjective report quality 7.5/10 &
Multimodal LLM systems can effectively generate radiology reports and support clinical decision-making &
Lower performance on complex pathologies; tooth numbering excluded; single LLM tested due to computational constraints \\ \addlinespace[6pt]

\cite{Balel2026} &
Improve radiology reporting using a hybrid framework integrating image analysis and LLMs &
30,954 panoramic radiographs; 50 unseen radiographs for evaluation &
YOLO-based detection and segmentation integrated with multiple LLMs using structured outputs &
DeepSeek R1 highest accuracy (466 true findings, 30 hallucinations); Gemma 3 highest hallucinations (60); commercial LLMs (ChatGPT-5, Gemini 2.5 Pro) hallucinated in 100\% of reports &
Hybrid models significantly improve reporting accuracy and reduce hallucinations compared to standalone LLMs &
Low detection of calculus/periodontal disease; segmentation errors propagate to reports; panoramic radiographs only\\ \addlinespace[6pt]

\cite{Stephan2025} &
Evaluate simplified AI-generated radiology reports for improving patient understanding &
3 versions of AI-generated radiology reports (original, simplified, optimized); evaluated by 300 patients &
FRE and LIX readability indices; 5-point Likert questionnaire (11 questions) &
FRE improved significantly (51.1$\rightarrow$55.0$\rightarrow$56.4, $p{<}0.001$); text 3 best in clarity, tone, structure; patient understanding: text 3 (1.5) vs.\ text 1 (3.1) &
ChatGPT simplification improves patient comprehension and engagement; clinical accuracy and generalizability need further research &
Simplified texts not evaluated for completeness; AI accountability unresolved; single-center study\\ \addlinespace[6pt]

\cite{Maruska2025} &
Compare ChatGPT and dentist responses for quality and empathy in dental patient queries &
20 patient questions randomly sampled from Reddit's dental advice community &
9 blinded dentists rated quality and empathy using 5-point Likert scales; best response selection &
ChatGPT responses were preferred in 93.3\% of cases and rated higher in both quality and empathy compared to dentist responses &
LLM chatbot responses show higher quality and empathy than online dentist responses; potential to enhance patient communication &
Online forum may not reflect real clinical communication\\ \addlinespace[6pt]

\cite{Alnsour2025} &
Evaluate ChatGPT-3.5 responses to patient inquiries on dental crown restorations&
126 patient questions categorized into three domains &
Usefulness rating; FKGL and SMOG readability; GQS quality; CLEAR reliability; PEMAT understandability and actionability &
Most responses were rated as useful or very useful (97.6\%); high quality (GQS mean 4.70/5) and reliability (CLEAR mean 24.37/25); readability remained difficult in many responses &
ChatGPT can serve as a supplementary patient education tool &
Single AI model tested; no layperson evaluation \\

\bottomrule
\end{tabular}
\end{adjustbox}
\end{table}

\section{Discriminative Vision Foundation Models}\label{sec:vision}
Where the models reviewed in Section~\ref{sec:gllm} are designed to produce
language, the models examined here learn structured visual representations:
embeddings, segmentation masks, or bounding boxes. In dentistry, three
architectures have been most prominently explored: SAM for prompt-based
segmentation~\cite{kirillov2023segment}, CLIP for contrastive
image-text alignment~\cite{radford2021learning}, and GroundingDINO for
open-vocabulary object detection~\cite{liu2024groundingdino}. These models share a common challenge: none were trained on dental data, and all require meaningful adaptation before they perform adequately on clinical dental imaging tasks, as reflected in the studies summarized in Table~\ref{tab:llm_vision_multimodal}.

\subsection{Tooth and Anatomical Segmentation}
SAM is the most extensively adapted foundation model in dental imaging. A consistent tradeoff emerges: changes that focus on a single limitation of SAM remain simpler but narrower in scope, whereas multi-stage pipelines that embed SAM within a larger system often achieve higher accuracy. However, their added complexity means that an error introduced at one stage can affect later stages.

Zero-shot SAM performance on dental images is generally weak. This is largely because dental radiographs differ substantially from the natural images used to train SAM. Key challenges include low contrast between tooth structures and surrounding bone, overlapping anatomies in panoramic views, fine boundary details at enamel–pulp interfaces, and the compression of three-dimensional structures into two-dimensional projections~\cite{Zhang2025EASAM,Wang2025ToothSegmentation,liao2024ppasam}.

 Several approaches address one specific limitation of SAM through a focused modification to its architecture. Boundary imprecision, for instance, has been addressed by augmenting SAM's encoder features with edge information from a parallel CNN branch, as proposed in EASAM. This improves tooth segmentation performance on panoramic dental X-rays through a parameter-efficient adaptation strategy~\cite{Zhang2025EASAM}.
 
 CBCT-specific robustness has been addressed differently by integrating SAM with a 3D VNet and adversarial learning in a dual-encoder architecture, as in PPA-SAM. This approach is intended to improve generalization across different CBCT imaging conditions~\cite{liao2024ppasam}. A broader version of this idea is used in Tooth-ASAM, which introduces adapter modules into SAM's encoder and mask decoder. These modules allow the model to adapt across CBCT scans, panoramic radiographs, and natural tooth images. This generally improved Dice, IoU, HD95, and ASSD compared with baseline SAM across the evaluated modalities~\cite{Wang2025ToothSegmentation}.

Other approaches embed SAM within a larger pipeline rather than modifying it directly, generally achieving stronger results at the cost of added complexity. Reducing annotation cost has been the goal of a two-stage framework, ToothSC-SAM, which combines region-of-interest extraction with prompt-based 3D SAM segmentation and skip connections to cut annotation time from several hours to minutes. It reaches approximately 93\% of fully supervised performance, suggesting that annotation cost, rather than peak accuracy, was the primary design objective~\cite{li2025toothscsam}.

Extending segmentation into three dimensions has been addressed by projecting 3D dental mesh data into 2D views for SAM processing and then reconstructing the result in 3D. This strategy is adopted by 3DTeethSAM, which achieved a mean IoU of 91.90\% on the 3DTeethSeg benchmark~\cite{lu2026_3dteethsam}. This reconstruction step introduces a dependence on accurate 3D-to-2D projection.

Splitting a hard visual task into smaller, easier sub-tasks is one way to improve performance, and the clearest demonstration of this comes from a pipeline combining detection, segmentation, and refinement into a single system. Tooth localization, initial mask generation, and boundary refinement together raised the Dice score from 0.672 for raw SAM to 0.9026~\cite{atchibay2026sam}. The detection stage identifies the tooth regions before segmentation, and the refinement stage corrects remaining boundary errors, so the segmentation depends on the quality of the detection stage that precedes it.

\begin{figure}[ht]
    \centering
    \includegraphics[width=\linewidth]{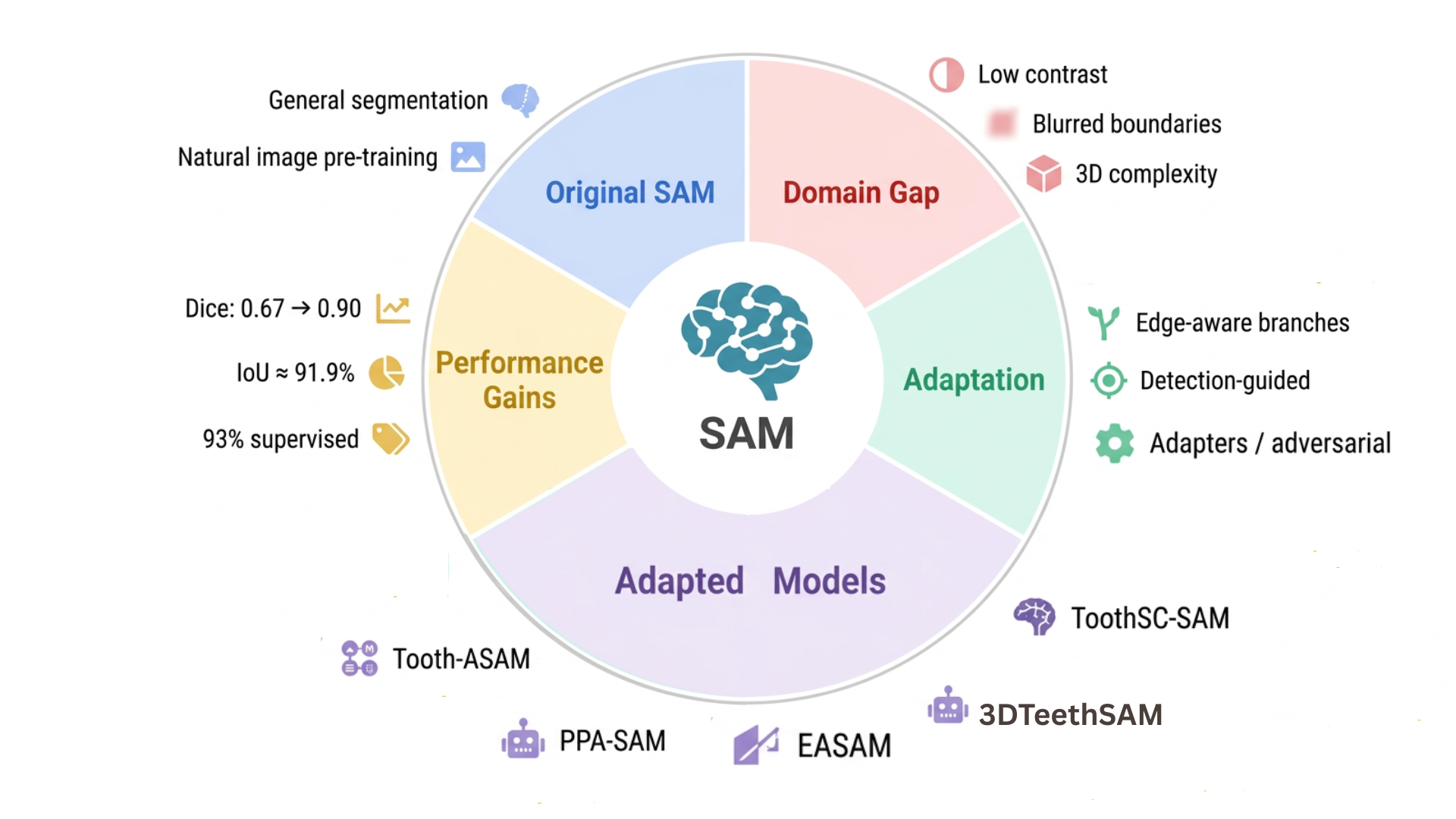}
    \caption{SAM adaptation strategies and adapted models for dental image
    segmentation, illustrating the range of architectural approaches
    used to address domain shift from natural image pretraining to
    clinical dental imaging.}
    \label{fig:sam_adaptation}
\end{figure}

The diversity of these strategies suggests that adaptation can occur either within the model architecture or through a larger pipeline. However, neither has emerged as a standard approach. Instead, the choice depends on the imaging modality, deployment constraints, and available resources, as summarized in Figure~\ref{fig:sam_adaptation}.

\subsection{Detection and Diagnosis}
Beyond segmentation, discriminative vision foundation models have been used for two  further tasks in dental imaging: detection, which locates where an abnormality is, and classification, which assigns a single label to the whole image. Detection is the harder of the two, since it must decide both whether an abnormality is present and where it lies, while classification only needs the overall label.

For both tasks, a general discriminative model does not transfer to dental images on its own, and the way it is adapted depends on what the task demands. Detection, being spatial, is improved by building dental structure into the detection process itself, so the model is guided toward the regions that matter. Classification, being image-level, is improved instead by retraining the visual representation on dental data, so the features themselves become dental.

This is what the reported work shows. For detection, a GroundingDINO-based framework guided by FDI tooth notation and the symmetry of the oral cavity reached 37.0\% mAP and 66.3\% AP50 on dental abnormality detection~\cite{Du2024DentalVLM}. For classification, a CLIP backbone fine-tuned on dental panoramic radiographs reached 0.929 accuracy and F1 on a four-category TMJOA task, while the unmodified and biomedical CLIP variants stayed below 0.40~\cite{kim2025dentalclip}. The failure of 
even the biomedical variants is the telling part: medical pretraining alone does not transfer, so the representation has to be specifically dental.

\begin{table}[!htbp]
\centering
\tiny
\setlength{\tabcolsep}{3pt}
\caption{Summary of studies evaluating Discriminative Vision Foundation Models in dentistry.}
\label{tab:llm_vision_multimodal}
\begin{adjustbox}{max width=\linewidth}
\begin{tabular}{
  >{\centering\arraybackslash}p{0.4cm}
  >{\RaggedRight}p{2.0cm}
  >{\RaggedRight}p{1.9cm}
  >{\RaggedRight}p{2.1cm}
  >{\RaggedRight}p{2.3cm}
  >{\RaggedRight}p{2.3cm}
  >{\RaggedRight}p{2.3cm}
}
\toprule
\textbf{Ref.} & \textbf{Objective} & \textbf{Dataset} & \textbf{Methods} & \textbf{Key Results} & \textbf{Conclusion} & \textbf{Limitations} \\
\midrule

\cite{Zhang2025EASAM} &
To improve tooth segmentation using edge-aware SAM-based architecture &
Tufts Dental Database (1,000 panoramic X-rays); generalization on BUSI and TN3K &
Dual-stream SAM encoder for salient features + CNN edge branch; feature fusion for segmentation &
Dice = 89.03\%, IoU = 80.56\%, HD = 40.84\% on tooth segmentation &
Edge-aware fusion enhances segmentation in dental X-ray images &
Fails when tooth contrast is similar to background; high parameter count increases computational cost \\ \addlinespace[6pt]

\cite{Wang2025ToothSegmentation} &
To adapt SAM for multimodal tooth segmentation &
NC CBCT (4,938 scans), private CBCT (45 volumes), MICCAI-Tooth (2,000/500), Vident-lab video dataset &
Adapter-based SAM (Tooth-ASAM) with modified encoder and mask decoder for dental images &
Improved Dice, IoU, HD95, and ASSD compared to baseline methods across all modalities &
SAM adaptation improves segmentation across multiple modalities &
Tooth adhesion, edge blurring, and small tooth segmentation unsolved; high computational cost \\ \addlinespace[6pt]

\cite{liao2024ppasam} &
To improve 3D tooth segmentation in CBCT using hybrid SAM-based architecture &
CBCT dataset, Hangzhou Dental Hospital (46 samples; 5, 10, 15, 20 training splits) &
VNet + adapter-finetuned SAM dual-encoder; three-layer convolutional discriminator for adversarial training &
Dice = 94.63\%, HD95 = 0.854, ASD = 0.207, Jaccard = 90.18\% at 20 training samples &
PPA-SAM outperforms other networks in accuracy and robustness for 3D CBCT tooth segmentation &
Training instability due to generator-discriminator imbalance; requires considerable samples, limiting zero-shot or one-shot performance \\ \addlinespace[6pt]

\cite{li2025toothscsam} &
To reduce annotation requirements for CBCT tooth segmentation &
176 CBCT scans, Sun Yat-sen University Affiliated Stomatological Hospital (2019--2022) &
Two-stage framework: ROI extraction followed by prompt-based 3D SAM with skip connections &
Dice = 83.88\%; over 90\% of fully supervised performance; annotation reduced from hours to minutes &
SAM-based prompting significantly reduces annotation effort while maintaining clinical-grade accuracy &
Single-center data with limited samples; still requires manual annotation, imposing a clinical burden \\ \addlinespace[6pt]

\cite{lu2026_3dteethsam} &
To adapt SAM2 for 3D tooth segmentation in high-resolution dental models &
Teeth3DS benchmark (1,800 intraoral 3D scans, 900 patients; 1,200 train / 600 test) &
Multi-view 2D rendering of 3D dental mesh; SAM2 segmentation with automatic prompt generator, mask refiner, and tooth classifier; 2D-to-3D reconstruction via voting strategy&
Overall accuracy = 95.48\%, mean tooth IoU = 91.90\%, boundary IoU = 70.05\%, Dice = 94.33\%; outperforms all compared methods including prior best method by +1.74\% tooth IoU  &
3DTeethSAM demonstrates the effectiveness of adapting 2D foundation models for 3D teeth segmentation, surpassing existing methods on the Teeth3DS benchmark &
2D-3D dimensionality mismatch requires multi-stage processing; performance dependent on multi-view rendering quality; limited by scarcity of labeled 3D dental data \\ \addlinespace[8pt]

\cite{atchibay2026sam} &
To improve SAM for tooth segmentation in panoramic radiographs &
Private dataset (224 images, Boston University) + public UFBA-UESC subset (425 images) &
YOLOv11 for tooth localization, SAM for initial segmentation, U-Net refinement for boundary correction &
YOLOv11 mAP@0.5 = 0.9931; Dice improved from 0.672 to 0.9026 after full pipeline &
Detection-guided prompting and refinement significantly improve SAM segmentation &
Sensitive to YOLO detection quality; union mask only, limiting tooth indexing and pathology localization \\ \addlinespace[6pt]

\cite{Du2024DentalVLM} &
To detect fine-grained dental abnormalities using vision-language models with dental notation-aware prompting &
DENTEX quadrant-enumerated subset (645 panoramic X-rays: 405 train, 102 val, 127 test) &
GroundingDINO with FDI notation prompts; multi-level detection (global and local) &
37.0\% mAP, 66.3\% AP50; +4.3\% mAP and +10.8\% AP50 over GroundingDINO baseline &
Dental notation-aware prompting improves fine-grained abnormality detection in panoramic X-rays &
Requires multiple fine-tuning stages; performance sensitive to prompt quality \\ \addlinespace[8pt]

\cite{kim2025dentalclip} &
To improve TMJOA diagnosis using contrastive vision-language learning &
Institutional PR dataset, Seoul National University Dental Hospital (446 labeled, 893 pseudo-labeled, 446 val, 446 test) &
CLIP (ViT-L/14) fine-tuned via contrastive learning with pseudo-labeling for TMJOA diagnosis &
Accuracy = 0.929 $\pm$ 0.012, F1-score = 0.929 $\pm$ 0.012; outperforms vanilla CLIP, BiomedCLIP, MedCLIP &
Contrastive semi-supervised learning improves TMJOA diagnosis under limited labeled data &
Requires paired CBCT and PR data; limited labeled data; generalization to other dental conditions not evaluated \\

\bottomrule
\end{tabular}
\end{adjustbox}
\end{table}

\section{Dental-Specific Foundation Models}\label{sec:dental}

The models in Sections~\ref{sec:gllm} and~\ref{sec:vision} are general-purpose
systems adapted to dental tasks. This section turns to models built around
dental knowledge from the start. Some are pretrained on large dental datasets.
Others are fine-tuned from general models on curated dental data. A third group
adds dental knowledge from outside the model, through ontologies, retrieval, or
validation steps. The main difference between these routes is whether dental
competence comes from data or from added structure. Section~\ref{sec:background}
introduced this distinction, and Table~\ref{tab:dental_foundation_models}
summarizes the studies discussed here.

\subsection{Pretrained-from-Scratch: DentVFM}

DentVFM is the only model in the reviewed literature that is a dental foundation
model in the architectural sense. It is trained from scratch on a large dental
imaging dataset, rather than adapted from a general-purpose
checkpoint~\cite{huang2025dentvfm}. It uses a Vision Transformer backbone in 2D
and 3D variants. Training is self-supervised, on DentVista, a dataset of about
1.6 million dental radiographic images. These span panoramic radiographs,
intraoral X-rays, anteroposterior and lateral X-rays, MRI, CT, and CBCT from
several medical centers. Because the objective is self-supervised, the model
learns dental visual features without large-scale manual annotation.

The model is evaluated on DentBench across a wide range of tasks. It outperforms
supervised, self-supervised, and weakly supervised baselines, and reaches
competitive performance with as little as 25\% labeled data. On some tasks, such
as cyst diagnosis and TMJ abnormality detection, it matches or exceeds
experienced dentists. It also shows cross-modality diagnosis, performing well
even when some imaging modalities are missing. This label efficiency and breadth
make it well suited to settings where annotated data and specialist coverage are
limited.

\subsection{Fine-Tuned Dental Models}

Most dental-specific systems are not trained from scratch. They are adapted from
general VLMs through instruction tuning, reinforcement learning, or multi-stage
training on curated dental data. Because large annotated dental corpora are hard
to assemble, the systems differ mainly in how they obtain supervision and how
their training signal is built.

The first point is how each system handles limited data, and this matters more than the architecture. The architecture is largely shared: both use two-stage training on the QwenVL-7B backbone family~\cite{meng2025dentvlm,zhang2025oralgpt}. What differs is how the image-language training signal is obtained. When a large curated corpus is available, the model can learn directly from dental image-question-answer pairs, as in DentVLM~\cite{meng2025dentvlm}. Where such data are unavailable, the signal is expanded from limited annotations: OralGPT uses similarity-guided pseudo-captioning to transfer descriptions from expert-annotated to weakly labeled images, allowing those images to support caption training~\cite{zhang2025oralgpt}. Under scarcity, the decisive design choice is how enough image-text supervision is obtained.

The second point is the training signal, and this has changed over time. Early
systems were narrow and focused on a single domain. CephGPT-4, for instance,
fine-tunes general vision-language models on cephalometric images and clinical
dialogue, with the landmark measurements supplied separately by a U-Net and
Steiner analysis~\cite{ma2023cephgpt4}. Later systems broadened the task and changed the signal. Reinforcement learning is added beyond plain instruction tuning, as in DentalGPT and OralGPT-Omni, on the idea that diagnosis needs step-by-step inference rather than simple recognition~\cite{cai2025dentalgpt,hao2025oralgptomni}. The most recent step is
agentic. OralGPT-Plus moves beyond single-pass inference by using an
inspect-zoom-compare loop that follows how a clinician re-reads a panoramic
film~\cite{fan2026oralgptplus}. Across these systems, compact 7B models trained
with reasoning supervision often outperform much larger
general-purpose models on dental tasks, which points to the training signal
rather than model size as the main driver.

Across these models, the field shows a clear need for dental adaptation rather than direct reuse of general-purpose VLMs. Once adapted with dental data and reasoning signals, compact models begin to outperform their own baselines and even compete with larger systems: DentalGPT rises from 46.7\% to 67.1\% average accuracy over its Qwen2.5-VL-7B-Instruct backbone, OralGPT-Omni outperforms GPT-5 on MMOral-Uni and MMOral-OPG, and DentVLM exceeds junior dentists on 21 of 36 tasks and senior dentists on 12 of 36 tasks. This also explains the recent move toward agentic designs, where the goal is to make the model examine dental images more like a clinician rather than produce a single static answer.

\subsection{LLM and VLM Driven Tools and Systems}
A further approach assembles dental capability from a pipeline rather than
concentrating it in a single trained model. An existing model is combined with
explicit structure, such as an ontology,
a retrieval index, or a validation step, which supplies much of the dental
specificity. Some pipelines also add a component of
their own, such as a task-specific segmentation tool or a fine-tuning step for
institutional phrasing, so the dental knowledge is spread across the system
rather than held in one place. The shared move is to lean on explicit structure
instead of large-scale dental training. This narrows the scope, but makes the
outputs easier to audit and to deploy. The systems differ mainly in the form the
structure takes.

The clearest case relies on a fixed ontology. In ArchMap, a hierarchical Dental Knowledge Base and a constrained schema turn a vision-language model into a
structured predictor for tooth counting, dentition-stage classification, and
conditions such as missing teeth, prosthetics, and caries, in a training-free
pipeline~\cite{Zhang2025ArchMap}. A second case relies on a validation gate. In
GumAgent, a vision-language model validates incoming images as intraoral,
denture, or non-relevant, so the trained segmentation model in its gum-disease
detection workflow only receives suitable photographs~\cite{Cheng2025GumAgent}.
A third case relies on retrieval. In ClinicGPT, an embedding-indexed institutional knowledge base supplies institution-specific context for administrative and protocol questions, while the model is fine-tuned to match institution-specific phrasing~\cite{Zhou2024ClinicGPT}. Their aim is narrower and more practical: dependable performance on a well-defined task, whether structured understanding of a scan, input validation for a detection pipeline, or administrative support, with reliability and deployability taking priority over diagnostic breadth.

\subsection{Datasets and Benchmarks}

The systems above have exposed a shared gap: the field still lacks large,
well-curated datasets and standard benchmarks. Two recent resources begin to
fill it. For panoramic analysis, MMOral pairs 20,563 annotated images with about
1.3 million instruction-following instances. These cover attribute extraction,
report generation, visual question answering, and image-grounded dialogue, and
evaluate models across clinical dimensions such as teeth condition, historical treatments, pathological
findings, and clinical recommendations~\cite{hao2025mmoral}. Its results show the domain gap clearly: GPT-4o reaches only a 41.45\% average score on MMOral-Bench, while one epoch of supervised fine-tuning on MMOral instruction data raises the Qwen2.5-VL-7B-based OralGPT from 21.46\% to 46.19\%, a gain of 24.73 percentage points~\cite{hao2025mmoral}. For dental
radiology more broadly, DentBench adds breadth, covering eight subspecialties,
more than 40 diseases, and seven modalities from 15 global
regions~\cite{huang2025dentvfm}. These resources are a starting point, and
broader shared benchmarks would help compare systems across studies.
\begin{table}[!htbp]
\centering
\scriptsize
\setlength{\tabcolsep}{4pt}
\caption{Summary of studies evaluating Dental Vision and Multimodal Foundation Models.}
\label{tab:dental_foundation_models}
\begin{adjustbox}{max width=\linewidth}
\begin{tabular}{
  >{\centering\arraybackslash}p{0.5cm}
  >{\RaggedRight}p{2.8cm}
  >{\RaggedRight}p{2.6cm}
  >{\RaggedRight}p{3.0cm}
  >{\RaggedRight}p{3.2cm}
  >{\RaggedRight}p{3.2cm}
}
\toprule
\textbf{Ref.} & \textbf{Objective} & \textbf{Dataset} & \textbf{Methods} & \textbf{Key Results} & \textbf{Conclusion} \\
\midrule

\cite{huang2025dentvfm} &
To develop a dental vision foundation model for generalizable dental imaging tasks &
DentVista dataset (~1.6M multimodal radiographic images) &
Vision Transformer-based model trained using self-supervised learning across multiple modalities &
Outperformed all 11 baselines; matched full-data performance with 25\% labeled data; exceeded dentists by +3.3\% (cyst diagnosis) and +13\% (TMJ detection)&
DentVFM enables scalable, generalizable dental image analysis \\ \addlinespace[4pt]

\cite{meng2025dentvlm} &
To develop a multimodal VLM for comprehensive dental diagnosis across 36 tasks and 7 modalities &
110,447 images from 20,741 patients, 31 provinces in China (2017--2024); bilingual; ~2.46M VQA pairs &
Two-stage training on Qwen2-VL-7B: vision-language alignment then diagnostic instruction tuning &
Outperformed junior dentists in 21/36 tasks, senior in 12/36; diagnostic time reduced 15--22\%; disease localization IoU = 38.44\% (EN), surpassing GPT-4o by 6.85\% &
DentVLM supports multimodal clinical decision-making for oral disease and malocclusion diagnosis \\ \addlinespace[4pt]

\cite{zhang2025oralgpt} &
To develop a VLM for oral mucosal disease diagnosis and description under low supervision &
480 expert-annotated + 659 weakly-labeled private + 1,280 public images; 4 conditions &
Two-stage training on Qwen2.5-VL-7B-Instruct: classification then caption generation with similarity-guided pseudo-captioning &
Classification F1 = 78.13\%, accuracy = 77.93\%; caption semantic avg = 6.36; BLEU-4 = 0.3736, METEOR = 0.5953 &
OralGPT supports diagnosis and clinical description generation under limited annotation \\ \addlinespace[10pt]

\cite{ma2023cephgpt4} &
To develop a multimodal cephalometric diagnostic system &
MD-QA (59,642 doctor--patient dialogues) + OCIMM cephalometric image dataset &
Fine-tuning Minigpt-4 and VisualGLM with cephalometric datasets and automated landmark detection &
Demonstrated strong performance in cephalometric analysis and diagnostic dialogue generation &
CephGPT-4 supports orthodontic measurement and diagnosis \\ \addlinespace[8pt]

\cite{cai2025dentalgpt} &
To improve multimodal reasoning for dental diagnosis using a specialized MLLM &
Over 120k dental images with detailed descriptions and QA pairs (PMC, open-source, and expert-annotated) &
Two-stage training on Qwen2.5-VL-7B-Instruct: multimodal understanding enhancement then GRPO-based reinforcement learning &
Average accuracy 67.1\% across 5 benchmarks, surpassing GPT-5 (59.2\%) and models with 100B+ parameters &
DentalGPT demonstrates that domain-specific data and reasoning training outperform model scale in dental diagnosis \\ \addlinespace[8pt]

\cite{hao2025oralgptomni} &
To develop a dental-specialized MLLM for comprehensive multimodal imaging analysis &
31 public + 1 in-house datasets; 8 modalities; 10+ countries &
Four-stage training on Qwen2.5-VL-7B: TRACE-CoT SFT then GRPO reinforcement learning &
MMOral-Uni = 51.84 (GPT-5 = 15.42); MMOral-OPG = 45.31; best among 27 MLLMs &
Enables transparent dental diagnosis via explicit chain-of-thought reasoning \\ \addlinespace[4pt]

\cite{fan2026oralgptplus} &
To develop an agentic VLM for panoramic radiograph analysis &
DentalProbe (5,000 images + expert trajectories); MMOral-X (300 QA pairs, 3 difficulty levels) &
Qwen2.5-VL-7B with Zoom-In and Mirror-In tools; instruction tuning then GRPO-based RL &
MMOral-X: Simple=43.16, Moderate=20.60, Complex=24.96; MMOral-OPG=45.35; best among all baselines &
Agentic symmetry-aware reasoning outperforms single-pass VLMs in panoramic diagnosis \\ \addlinespace[6pt]

\cite{Zhang2025ArchMap} &
To develop a knowledge-guided VLM system for structured dental understanding from 3D scans &
1,060 pre/post-orthodontic STL cases (435 subjects) &
Training-free pipeline combining geometry processing and ontology-guided reasoning &
Achieved improved accuracy and stability compared to supervised and prompted VLM approaches &
ArchMap enables structured dental understanding without training \\ \addlinespace[4pt]

\cite{Cheng2025GumAgent} &
To develop a VLM-based system for gum disease detection and image validation &
VLM: 400 mouth + 400 denture + 400 non-relevant; Segmentation: 332/83 train/test  &
Multi-stage pipeline including VLM classification and segmentation-based disease detection &
VLM validation accuracy = 88.58\%; gum disease segmentation Dice = 0.619, IoU = 0.482 &
GumAgent enables accessible gum disease detection via VLM-guided image validation \\ \addlinespace[4pt]

\cite{Zhou2024ClinicGPT} &
To develop an LLM-based system for improving dental clinic workflow efficiency &
Clinical workflow and institutional data &
Retrieval-based system integrating knowledge base with language model &
Demonstrated potential to improve clinical workflow efficiency and information access &
ClinicGPT supports clinical workflow and administrative tasks \\

\bottomrule
\end{tabular}
\end{adjustbox}
\end{table}

\section{Discussion, Limitations, and Future Directions}

\begin{figure}[ht]
\centering
\includegraphics[width=\textwidth]{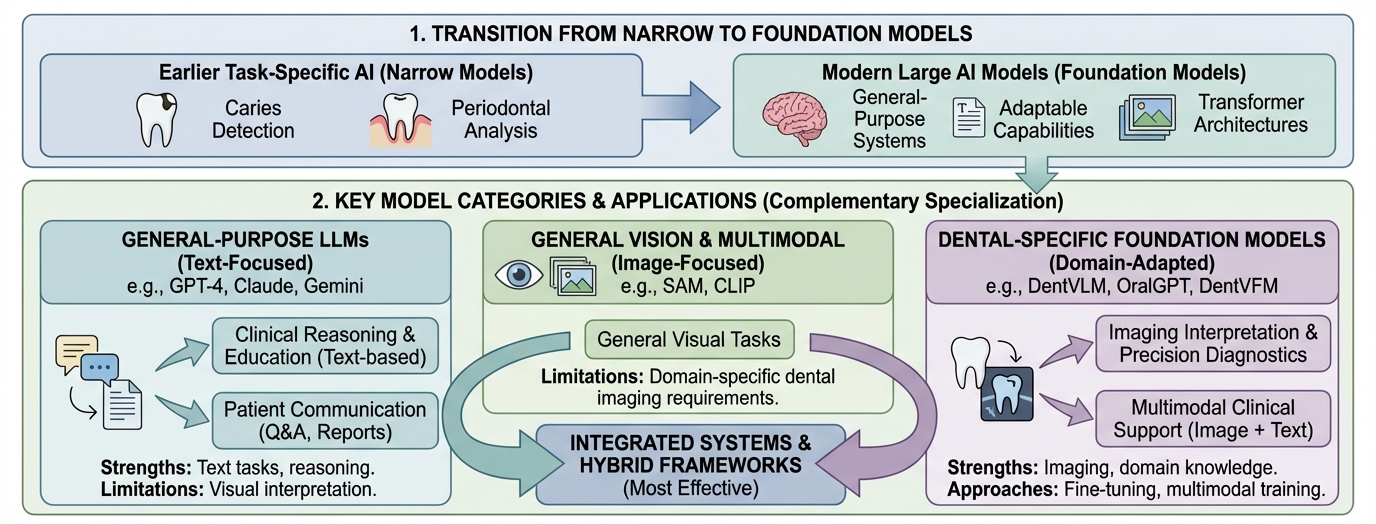}
\caption{Transition from narrow to foundation models and key model
categories with their complementary roles in dental healthcare.}
\label{fig:transition_taxonomy}
\end{figure}

\subsection{Comparison with Existing Reviews}

The contribution of this review lies not in the number of studies it covers
but in examining all three model families together. Prior reviews have each
addressed only a subset; none has examined language-generative, discriminative
vision, and dental-specific foundation models within a single framework. The
transition from narrow task-specific systems to integrated foundation-model
ecosystems is illustrated in Figure~\ref{fig:transition_taxonomy}.

The reviews closest to this one are confined to the language-generative family,
and the present review both confirms their findings and updates them. Umer et
al.~\cite{umer2024llm_dentistry_scoping} (BDJ Open, 2024) identified 17
ChatGPT-dominated studies, with Likert scales the most common metric and
advanced prompting in only two. The 97 studies surveyed here through early 2026
mark a phase transition rather than a wider search: model diversity now spans
Gemini, Claude, DeepSeek, and domain-specific systems, and advanced prompting
and retrieval-augmented generation, rare in earlier reviews, now recur across
many studies. Within education, Aura-Tormos et
al.~\cite{auratormos2025dental_education} reviewed 60 studies and reported that
GPT-4 outperformed GPT-3.5, that curricular integration remained informal, and
that misinformation and overreliance were common. The present review confirms
these and adds two findings absent from that account: a text--image performance
divide across examination systems, and temporal instability of outputs. Within
oral and maxillofacial surgery, Ronsivalle et al.~\cite{ronsivalle2025omfs}
reached a similar conclusion of constrained performance in complex, emotionally
sensitive scenarios from only four studies, which the present review grounds in
a much larger evidence base.

More broadly, Busch et al.~\cite{busch2025patient_care} reviewed 89 studies
across 29 specialties, identifying design limitations (lack of domain
optimization, data transparency) and output (non-reproducibility,
incorrectness) as the two primary domains. The dental findings map directly onto this: absent
domain optimization corresponds to the image-dependent performance gap, and
output limitations such as hallucination and inconsistency recur across the
clinical reasoning and patient communication sections. What dentistry adds is
that the text--image divide is sharper and more consequential here than in
most specialties, given the centrality of radiographic interpretation.

For discriminative vision models no comparable dental review exists. The nearest
is Ryu et al.~\cite{ryu2025vlm_medical}, who reviewed medical vision-language
models from 2022 to 2024 and reported promise alongside generalization and
integration challenges, both replicated in the dental evidence here. To this the
present review adds that no shared evaluation benchmark yet exists for comparing
discriminative vision models in dentistry, which is the single largest obstacle
to assessing them against one another.

\subsection{Complementarity and the Dimensions of Clinical Deployment}

\begin{sidewaysfigure}
    \centering
    \includegraphics[width=\linewidth]{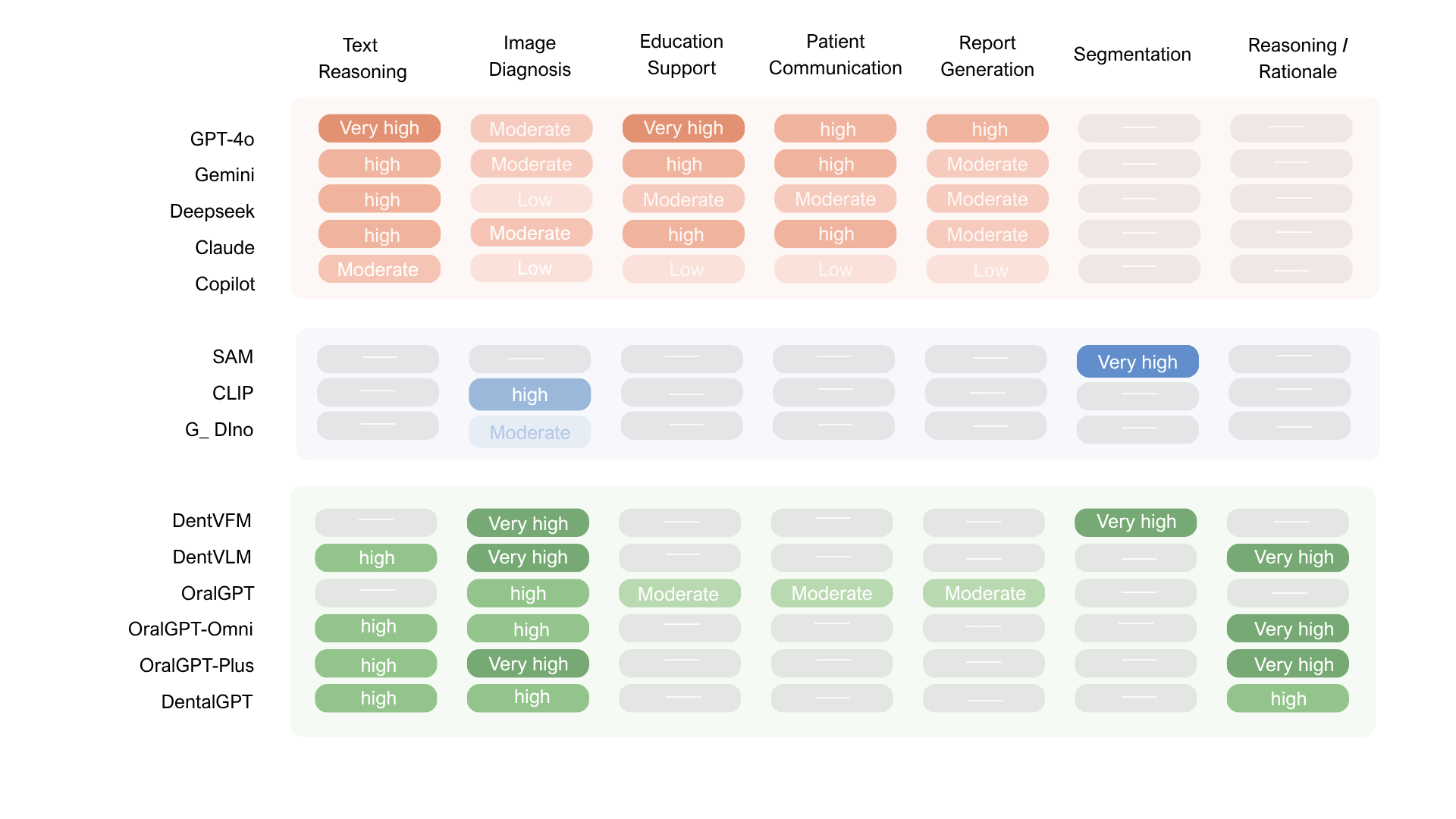}
    \caption{Performance heatmap of large AI models across dental task
    categories. Ratings (Moderate, High, Very High) reflect a qualitative
    synthesis by the authors based on the range of quantitative performance
    metrics reported across included studies for each model--task combination;
    they are not derived from a single threshold but represent the authors'
    assessment of the overall evidence. Empty cells indicate the task was not
    evaluated or reported for that model.}
    \label{fig:heatmap}
\end{sidewaysfigure}

The most practically significant finding of this review is that
language-generative models, discriminative vision models, and
dental-specific foundation models are complementary rather than competing,
each strongest precisely where the others are weakest. Language-generative
models excel at reasoning, education, and dialogue but fail on
image-dependent diagnosis. Discriminative vision models achieve spatial
precision in segmentation and detection but emit no language.
Dental-specific foundation models reach the strongest multimodal
performance but demand training data and compute that limit their
availability. As shown in Figure~\ref{fig:heatmap}, these profiles occupy
distinct regions of the task space, with no single category dominating all
columns.

This complementarity is reflected in the integrated-pipeline literature,
where combinations of model types consistently outperform their individual
components. Hybrid pipelines that combine discriminative vision models for
image interpretation with language models for report generation reduce
hallucination and improve report quality relative to standalone language
models~\cite{Dasanayaka2025,Balel2026}. Retrieval-augmented language models
improve performance on knowledge-intensive clinical questions by grounding
responses in external knowledge sources~\cite{Fanelli2025,Rewthamrongsris2026},
while agentic and knowledge-guided vision-language systems outperform
conventional single-pass approaches in panoramic radiograph interpretation
and 3D dental analysis~\cite{fan2026oralgptplus,Zhang2025ArchMap}. Taken
together, these findings suggest that performance improvements are often
achieved by combining complementary model capabilities rather than by
relying on a single model architecture.

Complementarity explains why these systems are combined, but not how they
differ once placed in a clinical setting. Four dimensions separate the
families and decide how each can be used: the optimization through which
dental competence is obtained, the way the clinician and the model
interact, the degree of decision-making autonomy each can be granted, and
the readiness of each for real-world deployment.

\textbf{Optimization.}
Most studies reach first for general-purpose generative models, for
practical rather than principled reasons: these models require the least
effort to deploy, carry broad knowledge already in their weights, and need
no dental data to produce a plausible answer. This works while the task
stays close to what the model already knows, and breaks down as the task
moves deeper into the specialty, where general knowledge thins out and
dental-specific visual or reasoning ability becomes decisive. How much
optimization a task needs, and which kind, is therefore set by the task
itself rather than chosen freely, and three properties of the task decide
it.
The first is how far the answer already sits in general knowledge. A
definitional or common-knowledge question, such as what dental caries is or
how a patient should care for a new crown, sits inside the base model and
needs no adaptation, so prompting alone is sufficient; a question that
depends on dental-specific visual detail, such as grading an impacted molar
or segmenting a tooth boundary, sits outside it and fails without
adaptation~\cite{Asar2025,kim2025dentalclip}.
The second property is whether
the required knowledge is stable or renewable. When the answer depends on
guidance that changes over time, such as treatment recommendations that change when clinical guidelines are updated, RAG is the better
choice than heavier optimization, because the knowledge base can be replaced
when the guideline changes without retraining, keeping the model
current~\cite{Rewthamrongsris2026,Fanelli2025}; committing the same
knowledge to the weights through fine-tuning is both more expensive and
quicker to go stale. When the required knowledge is stable, such as
anatomical structure or fixed diagnostic reasoning, fixing it in the weights
is the point, and fine-tuning or from-scratch training pays
off~\cite{huang2025dentvfm}. 
The third property is the form of the
difficulty: a task that needs a fixed output format or a worked pattern is
helped most by in-context examples~\cite{rewthamrongsris2025image,Gao2026LLM},
a task that needs several reasoning steps, such as reading a panoramic film,
is helped most by reasoning supervision or reinforcement
learning~\cite{Ji2025ChatGPTOralMaxillofacial,cai2025dentalgpt,hao2025oralgptomni},
and a task that needs spatial precision is helped only by adapting a vision
model on dental images~\cite{Wang2025ToothSegmentation}. 
The mechanism
therefore differs by family as well as by task: generative models are
optimized mainly through their input, by prompting and retrieval, whereas
discriminative vision models are optimized through their parameters, by
adaptation and fine-tuning. These choices are further bounded by factors
outside the task, since heavier optimization buys performance at the cost of
annotated data, compute, and maintenance, all of which are scarce in
dentistry. The practical rule that follows is to use the lightest
optimization the task allows, and to move up the tiers only when task
difficulty, the need for dental-specific knowledge, or the stability of that
knowledge demands it.

\textbf{Human-agent interaction.}
Where optimization concerns how a model acquires competence, interaction
concerns how the clinician and the model exchange information while the model
is used, and the families differ here as much as on any other axis. With
discriminative vision models the interaction is narrow and largely
one-directional: the clinician supplies a prompt, a point, a box, or a class
name, and then verifies a structured output that can be checked directly
against the image~\cite{kim2025dentalclip,Wang2025ToothSegmentation}. With
language-generative models the interaction is conversational and iterative,
allowing follow-up questions, reformulation, and clarification, which is more
flexible but also more demanding, because fluent output can hide missing or
incorrect content and the clinician must stay vigilant rather than simply
verify~\cite{Stephan2024}. The agentic dental models change this shape again:
where a clinician would normally drive the re-examination of a film step by
step, an inspect-zoom-compare loop internalizes part of that process and
exposes it as a reasoning trace~\cite{fan2026oralgptplus}, so the clinician's
role shifts from steering each step to auditing the steps the model took. The
same shift underlies the move toward chain-of-thought supervision, which is
valued less for raw accuracy than for making the model's reasoning visible
enough to check~\cite{hao2025oralgptomni}.

\textbf{Decision-making autonomy.}
The interaction a model supports also bounds how much it can be allowed to
decide. Discriminative vision foundation models produce structured outputs,
such as segmentation masks, landmark coordinates, and similarity scores, that
can be quantitatively evaluated and independently verified, which makes them
the most plausible candidates for limited autonomy in narrowly defined tasks
such as segmentation, measurement, and landmark detection, with the clinician
serving as a verifier. Language-generative models occupy a different
position: although they perform strongly on reasoning, communication, and
knowledge-intensive tasks, their outputs remain probabilistic and vulnerable
to hallucination, so their role is best understood as assistive rather than
autonomous, with clinician review essential regardless of benchmark
performance. Dental-specific foundation models fall between these extremes,
their stronger multimodal performance supporting more advanced decision
support, but human oversight remaining necessary.

\textbf{Deployment readiness.}
A mismatch separates current research practice from likely clinical use. Most
studies still evaluate models as isolated systems, yet the strongest results
emerge from architectures that combine several model types in one workflow,
and most dental-specific models have been tested on benchmark datasets rather
than in prospective clinical studies. Closing this gap calls for integration
frameworks, and evaluation protocols that
assess complete workflows rather than individual models, alongside the
prospective studies that benchmark accuracy cannot substitute for.

Read across these four dimensions, the field is moving in one direction: from
general models used off the shelf toward systems whose optimization is
matched to the task, whose reasoning is made explicit enough to audit, and
whose components are combined rather than used alone. Early work could treat a
general model as the whole system; the strongest recent work treats it as one
part, adapted where the task is specialized, retrieved into where the
knowledge is renewable, and wrapped in a reasoning or pipeline structure where
a single pass is not enough. The open question for the coming years is less
which single model is best, and more how deliberately these choices are
matched to the task and the constraints of cost, data, and time.

\subsection{Challenges, Research Gaps, and Future Directions}

The challenges identified in this review are closely linked to gaps in the current evidence. These gaps also indicate the areas where further research is most needed. Table~\ref{tab:challenges} brings together the major challenges, the reasons they remain unresolved, and the directions most likely to address them.

The following discussion examines the challenges that remain the greatest barriers to clinical deployment.

\begin{table*}[!t]
\centering
\scriptsize
\setlength{\tabcolsep}{4pt}
\caption{Challenges, research gaps, and corresponding future directions in large AI models for dental healthcare, organized by theme.}
\label{tab:challenges}

\begin{adjustbox}{max width=\textwidth}
\begin{tabular}{
>{\RaggedRight\arraybackslash}p{2.5cm}
>{\RaggedRight\arraybackslash}p{4.5cm}
>{\RaggedRight\arraybackslash}p{5.0cm}
>{\RaggedRight\arraybackslash}p{4.7cm}
}
\toprule
\textbf{Theme} &
\textbf{Challenge / Research gap} &
\textbf{Why it remains unresolved} &
\textbf{Future direction} \\
\midrule
\multirow{3}{*}{\parbox[t]{2.5cm}{Evaluation\\Benchmarking}}
&
No standardized clinical benchmark
&
Studies use different datasets, evaluation metrics, and scoring methods, making direct comparison difficult
&
Shared clinical benchmarks with fixed test sets and independent external validation
\\ \addlinespace[8pt]

&
Absence of external, cross-institution validation
&
Most models are evaluated using data from a single institution or patient population
&
Multi-center validation across institutions, imaging devices, and patient populations
\\ \addlinespace[8pt]

&
Limited real-world clinical evaluation
&
Most evidence comes from retrospective benchmark studies rather than clinical use
&
Prospective clinical studies evaluating performance in routine practice
\\

\midrule

\multirow{3}{*}{Data}
&
Scarcity of large, annotated dental datasets, especially text corpora
&
Expert annotation is costly; large-scale dental text resources are limited
&
Self-supervised and annotation-efficient pretraining; similarity-guided pseudo-captioning
\\ \addlinespace[8pt]

&
Unclear data provenance and quality control
&
Limited reporting of collection protocols, demographics, and annotation QC
&
Standardized dataset documentation and transparent provenance reporting
\\ \addlinespace[8pt]

&
Geographic and demographic bias (East Asia, Turkey, US concentration)
&
Datasets sourced from few regions; no data from low-income settings where oral-disease burden is highest
&
Geographically and demographically diverse datasets; validation in under-served settings
\\

\midrule

\multirow{3}{*}{\parbox[t]{2.5cm}{Reliability\\ \& Safety}}
&
Hallucination acknowledged but rarely quantified
&
No standard metric for hallucination on clinical dental tasks
&
Quantification frameworks reporting hallucination rate, type, and clinical severity
\\ \addlinespace[8pt]

&
Persistent text--image performance gap
&
Vision-language models remain less reliable on image-dependent tasks, particularly radiographic interpretation
&
Stronger visual grounding and calibrated confidence; hybrid pipelines that delegate localization to discriminative vision models
\\ \addlinespace[8pt]

&
Temporal instability of outputs across sessions
&
Model and version drift; nondeterministic decoding
&
Version-pinned evaluation and explicit reproducibility reporting
\\

\midrule

\multirow{2}{*}{\parbox[t]{2.5cm}{Clinical \&\\ Demographic\\ Coverage}}
&
Rare diseases, head-and-neck oncology, cleft lip/palate, geriatric and paediatric imaging underrepresented
&
Low prevalence, annotation scarcity, and research concentration on common, well-defined tasks
&
Targeted curation for underrepresented conditions and populations
\\ \addlinespace[8pt]

&
Limited integration across clinical data modalities
&
Difficulty assembling paired multimodal data; complexity of fusion and joint reasoning
&
Joint modeling of imaging, electronic health records, longitudinal history, and clinical notes
\\

\midrule

\multirow{2}{*}{\parbox[t]{2.5cm}{Deployment\\ \& Regulation}}
&
No engagement with regulatory pathways (FDA SaMD; EU AI Act high-risk classification)
&
Research-stage focus; limited regulatory expertise in the modeling literature
&
Regulatory-aware design: risk classification, transparency, human oversight, and post-market surveillance planning
\\ \addlinespace[8pt]

&
Lack of deployment-oriented autonomy and oversight frameworks
&
Single-model benchmark accuracy treated as the end goal; deployment considered out of scope
&
Defined autonomy levels and clinician oversight mechanisms for clinical deployment
\\

\bottomrule
\end{tabular}
\end{adjustbox}
\end{table*}

A major challenge in interpreting the existing literature is the lack of consistent evaluation standards. Differences in question format, scoring rubrics, evaluation metrics, and evaluator expertise make reported performance difficult to compare across studies. As a result, it is often unclear whether observed differences reflect genuine improvements in model capability or differences in how performance was assessed. This problem extends beyond benchmarking itself. Claims regarding reliability, generalizability, and clinical utility all depend on the validity of the underlying evaluation framework. Future progress therefore requires shared benchmarks, standardized evaluation protocols, and independent external validation capable of supporting meaningful comparison across studies.

Limitations in available data extend beyond sample size alone. The data problem in dental AI is not limited to small sample sizes. Large annotated corpora remain difficult to build because expert labeling is costly, large dental text corpora remain scarce, and many datasets are not shared openly. As a result, much dental-specific capability is still obtained through adaptation of general-purpose models rather than large-scale domain-specific pretraining. Existing datasets are also concentrated in a limited number of regions and institutions, with limited reporting of collection protocols, patient demographics, and annotation procedures. Future progress will require not only larger datasets, but also more representative, better documented, and more accessible data resources.

Reliability remains the principal barrier to the autonomous use of dental AI systems. Hallucination, instability across repeated interactions, and weaker performance on image-dependent tasks all reduce confidence in model outputs when clinical decisions depend on them. The persistent text--image performance gap is particularly important because it highlights the difficulty of transferring strong language capabilities to visual diagnostic tasks. Although hybrid systems partially address this limitation by separating image interpretation from language generation, reliability remains difficult to assess because these failures are rarely measured systematically. Future progress therefore depends not only on improving performance, but also on developing standardized methods for quantifying error frequency, severity, and clinical impact.

Coverage and deployment limitations constrain generalizability. The reviewed evidence is geographically concentrated, clinically narrow, and almost entirely single-modality, while engagement with the regulatory frameworks that will govern clinical deployment remains absent. None of the reviewed studies discusses the FDA Software as a Medical Device framework or the EU AI Act's high-risk provisions, despite the fact that many of the clinical decision-support applications reviewed in Sections~4 through~6 would likely fall within these categories. Considerations such as risk classification, transparency requirements, autonomy limits, and post-market surveillance are similarly absent from the literature. Future research should therefore move beyond benchmark evaluation toward clinically representative settings, broader patient populations, multimodal workflows, and deployment frameworks that explicitly define autonomy levels and human oversight.

\subsection{Limitations of This Review}
Several limitations should be acknowledged. The 2020 to early 2026 search window captures the foundation-model era but excludes earlier deep-learning work. Including arXiv preprints reflects the dental foundation-model literature but introduces non-peer-reviewed evidence. Although two reviewers (S.H. and R.D.) screened independently, inter-rater reliability was not calculated. The two-dimensional framework in Section~3 is theoretical and has not been empirically validated; it describes the literature rather than explains it. The English-language restriction may exclude relevant studies in other languages. Finally, heterogeneous outcome measures prevented meta-analysis, so comparisons remain narrative.

\section{Conclusion}

This review examined 97 studies from 2020 to 2026 through a two-dimensional
framework based on architectural paradigm and degree of dental specialization.
The clearest empirical pattern is a divide between text and image:
language-generative models perform strongly on text-based reasoning and
examination tasks but weaken sharply on image-dependent diagnosis, while
dental-specific foundation models reach the highest performance on complex
multimodal tasks at the cost of substantial data and compute.

Two structural lessons follow. Adapting a model to dental use is a tiered choice
rather than a search for one best method, since gains grow with how deeply the
model is changed and the lightest tiers cannot close the image gap; the
practical task is to match the strategy to the available data, compute, and
task. And no single model class covers the full range of clinical work, so the
strongest systems combine complementary types within structured pipelines. Both
point the same way: progress depends more on sound adaptation choices and
integration than on optimizing individual models against narrow benchmarks.

Safe autonomous deployment remains conditional on three unresolved barriers:
unquantified hallucination, scarce and unevenly distributed dental data, and the
absence of standardized clinical benchmarks. Until these are addressed, the
appropriate role for these systems is as supervised assistive tools that augment
clinical expertise. The most promising directions, including agentic reasoning,
annotation-efficient pretraining, and knowledge-guided pipelines, are advancing
quickly, but whether they reach clinical deployment will depend as much on
regulatory clarity, data infrastructure, and shared evaluation standards as on
further model innovation.

\begin{table}[!htbp]
\centering
\scriptsize
\setlength{\tabcolsep}{4pt}
\caption{List of abbreviations used in this review.}
\label{tab:abbreviations}
\begin{adjustbox}{max width=\linewidth}
\begin{tabular}{
  >{\bfseries\RaggedRight}p{1.8cm}
  >{\RaggedRight}p{5.7cm}
  >{\bfseries\RaggedRight}p{1.8cm}
  >{\RaggedRight}p{5.7cm}
}
\toprule
\textbf{Abbreviation} & \textbf{Full Term} & \textbf{Abbreviation} & \textbf{Full Term} \\
\midrule
\multicolumn{4}{l}{\textit{AI \& Machine Learning}} \\[4pt]
BERT  & Bidirectional Encoder Representations from Transformers &
CLIP  & Contrastive Language--Image Pretraining \\ \addlinespace[2pt]
CNN   & Convolutional Neural Network &
CoT   & Chain-of-Thought \\ \addlinespace[2pt]
DINO  & Detection Transformer with Improved DeNoising Anchor Boxes &
ICL   & In-Context Learning \\ \addlinespace[2pt]
LLM   & Large Language Model &
LVLM  & Large Vision--Language Model \\ \addlinespace[2pt]
MLLM  & Multimodal Large Language Model &
RAG   & Retrieval-Augmented Generation \\ \addlinespace[2pt]
RL    & Reinforcement Learning &
SAM   & Segment Anything Model \\ \addlinespace[2pt]
ViT   & Vision Transformer &
VLM   & Vision--Language Model \\ \addlinespace[2pt]
VQA   & Visual Question Answering &
      & \\ [8pt]
\multicolumn{4}{l}{\textit{Clinical \& Dental Terms}} \\[4pt]
CBCT  & Cone Beam Computed Tomography &
FDI   & F\'{e}d\'{e}ration Dentaire Internationale \\ \addlinespace[2pt]
IADT  & International Association of Dental Traumatology &
IAPD  & International Association of Paediatric Dentistry \\ \addlinespace[2pt]
OMFS  & Oral and Maxillofacial Surgery &
OPG   & Orthopantomogram \\ \addlinespace[2pt]
TMJOA & Temporomandibular Joint Osteoarthritis &
ITI   & International Team for Implantology \\ [8pt]
\multicolumn{4}{l}{\textit{Evaluation \& Metrics}} \\[4pt]
AP50    & Average Precision at IoU threshold 0.50 &
ASSD    & Average Symmetric Surface Distance \\ \addlinespace[2pt]
CLEAR   & Comprehensibility, Learnability, Empowerment, Actionability, Readability &
CVI     & Content Validity Index \\ \addlinespace[2pt]
DISCERN & Tool for assessing health information quality &
F1      & F1 Score (harmonic mean of precision and recall) \\ \addlinespace[2pt]
FKGL    & Flesch--Kincaid Grade Level &
FRE     & Flesch Reading Ease \\ \addlinespace[2pt]
FRES    & Flesch Reading Ease Score &
GQS     & Global Quality Scale \\ \addlinespace[2pt]
HD95    & 95th-percentile Hausdorff Distance &
ICC     & Intraclass Correlation Coefficient \\ \addlinespace[2pt]
IoU     & Intersection over Union &
LIX     & L\"{a}sbarhetsindex (Swedish readability index) \\ \addlinespace[2pt]
mAP     & Mean Average Precision &
MCQ     & Multiple-Choice Question \\ \addlinespace[2pt]
PEMAT   & Patient Education Materials Assessment Tool &
PRISMA  & Preferred Reporting Items for Systematic Reviews and Meta-Analyses \\ \addlinespace[2pt]
QA      & Question Answering &
SMOG    & Simple Measure of Gobbledygook \\ [8pt]
\multicolumn{4}{l}{\textit{Examinations \& Guidelines}} \\[4pt]
AAE   & American Association of Endodontists &
ADA   & American Dental Association \\ \addlinespace[2pt]
ADAT  & Advanced Dental Admission Test &
AHA   & American Heart Association \\ \addlinespace[2pt]
API   & Application Programming Interface &
DAT   & Dental Admission Test \\ \addlinespace[2pt]
DUS   & Di\c{s} Hekimli\u{g}i Uzmanl{\i}k S{\i}nav{\i} (Turkish Dentistry Specialization Examination) &
ESE   & European Society of Endodontology \\ \addlinespace[2pt]
INBDE & Integrated National Board Dental Examination &
JNDE  & Japanese National Dental Examination \\ \addlinespace[2pt]
KNDLE & Korean National Dental Licensing Examination &
NDLE  & National Dental Licensing Examination \\ \addlinespace[2pt]
AAO   & American Association of Orthodontists &
ACP   & American College of Prosthodontists \\ \addlinespace[2pt]
EAO   & European Association for Osseointegration &
      & \\
\bottomrule
\end{tabular}
\end{adjustbox}
\end{table}


\section*{CRediT Author Contribution Statement}
\textbf{Sema Helali:} Conceptualization, Methodology, Data curation,
Formal analysis, Investigation, Writing -- original draft, Writing --
review and editing.
\textbf{Lina Abu Nada:} Clinical expertise, Validation of clinical
relevance and implications of AI in dental practice, Writing --
review and editing.
\textbf{Alaa Abd-Alrazaq:} Methodology, Data extraction mechanism,
Writing -- review and editing (Discussion).
\textbf{Sausan Al Kawas:} Clinical expertise, Validation of clinical
relevance and implications of AI in dental practice, Writing --
review and editing.
\textbf{Faleh Tamimi:} Clinical expertise, Guidance of analysis, Writing --
review and editing.
\textbf{Rafat Damseh:} Conceptualization, Methodology, Data curation,
Formal analysis, Validation, Writing -- review and editing, Supervision,
Project administration.

\section*{Declaration of Competing Interest}
The authors declare that they have no known competing financial interests
or personal relationships that could have appeared to influence the work
reported in this paper.

\section*{Funding}
This work was supported by United Arab Emirates University through the
UAEU Strategic Research Grant 12R315.

\section*{Data Availability}
The data supporting the findings of this review are available from the
corresponding author upon reasonable request.

\section*{Ethics Approval}
This study is a scoping review of publicly available literature and did
not involve human participants, animal subjects, or identifiable personal
data. Ethical approval was therefore not required.


\end{document}